\documentclass[conference, compsoc]{IEEEtran}
\ifCLASSOPTIONcompsoc
\else
\fi

\usepackage[backend=bibtex,bibstyle=ieee,citestyle=numeric-comp]{biblatex}
\AtBeginBibliography{\small}
\bibliography{Bibliography}
\usepackage[utf8]{inputenc}
\usepackage[T1]{fontenc}
\usepackage[english]{babel}
\usepackage{eurosym}
\usepackage{url}
\usepackage{graphicx}
\usepackage{setspace}
\usepackage{float}
\usepackage{multirow, multicol}
\usepackage{amsmath}
\usepackage{amsfonts}
\usepackage{amssymb}
\usepackage{stmaryrd}
\usepackage{tikz}
\usepackage{tkz-euclide}
\usetikzlibrary{matrix,chains,positioning,decorations.pathreplacing,arrows, calc, fit}
\usepackage{stmaryrd}
\usepackage{fancybox}
\usepackage{xcolor}
\usepackage{hhline}
\usepackage{epsf}
\usepackage{rotating}
\usepackage{chngcntr}
\usepackage{setspace}
\usepackage{blindtext}
\usepackage{array}
\usepackage{listings}
\usepackage{colortbl}
\usepackage{verbatim}
\usepackage{mathrsfs}
\usepackage{enumerate}
\usepackage{subfiles}
\usepackage{wrapfig}
\usepackage{diagbox}
\usepackage{bbm}
\usepackage{amsthm}
\definecolor{dpgreen}{rgb}{0.01, 0.75, 0.24} 
\definecolor{ao}{rgb}{0.0, 0.5, 0.0} 
\usepackage[colorlinks=true,
pdfpagelabels,
pdfstartview = FitH,
bookmarksopen = true,
bookmarksnumbered = true,
linkcolor = dpgreen,
plainpages = false,
hypertexnames = false,
citecolor = dpgreen, 
urlcolor=black]{hyperref}
\usepackage{soul}
\usepackage{numprint}
\usepackage[normalem]{ulem}
\npdecimalsign{.}
\nprounddigits{3}

\usepackage{nicefrac}
\theoremstyle{plain}
\newtheorem{theorem}{Theorem}[section]

\newtheorem{remark}[theorem]{Remark}
\newtheorem{definition}[theorem]{Definition}

\usepackage{xpatch}
\makeatletter
\AtBeginDocument{\xpatchcmd{\@thm}{\thm@headpunct{.}}{\thm@headpunct{}}{}{}}
\makeatother

\usepackage[noabbrev]{cleveref}

\def\R{\mathbb{R}}
\def\F{\mathcal{F}}
\def\L{\mathcal{L}}
\def\x{\mathbf{x}}
\def\y{\mathbf{y}}

\definecolor{cblue}{rgb}{0.16, 0.32, 0.75} 
\definecolor{cobalt}{rgb}{0.0, 0.28, 0.67}
\definecolor{cyan_p}{rgb}{0.0, 0.72, 0.92} 
\definecolor{maroon_html}{rgb}{0.5, 0.0, 0.0}
\definecolor{pumpkin}{rgb}{1.0, 0.46, 0.09}
\definecolor{persimmon}{rgb}{0.93, 0.35, 0.0}
\definecolor{maroon_x11}{rgb}{0.69, 0.19, 0.38}

\definecolor{bblue}{rgb}{0.74, 0.83, 0.9} 
\definecolor{pasyel}{rgb}{0.99, 0.99, 0.59} 
\definecolor{blush}{rgb}{0.87, 0.36, 0.51} 
\definecolor{pistachio}{rgb}{0.58, 0.77, 0.45}
\definecolor{mauvelous}{rgb}{0.94, 0.6, 0.67}
\definecolor{cblue}{rgb}{0.6, 0.73, 0.89} 
\definecolor{colblue}{rgb}{0.61, 0.87, 1.0} 
\definecolor{navy}{rgb}{0.0, 0.0, 0.5}

\definecolor{lgray}{rgb}{0.75, 0.75, 0.75}
\definecolor{lgray1}{rgb}{0.83, 0.83, 0.83}
\definecolor{lgray2}{rgb}{0.9, 0.9, 0.9}
\definecolor{orange}{rgb}{0.98, 0.6, 0.01}
\definecolor{porange}{rgb}{1.0, 0.7, 0.28}
\definecolor{dorange}{rgb}{1.0, 0.55, 0.0}
\definecolor{orange-red}{rgb}{1.0, 0.27, 0.0}
\definecolor{deepcarrotorange}{rgb}{0.91, 0.41, 0.17}
\definecolor{dred}{rgb}{0.55, 0.0, 0.0}
\definecolor{pred}{rgb}{1.0, 0.41, 0.38}
\definecolor{egreen}{rgb}{0.0, 0.5, 0.0}
\definecolor{bred}{rgb}{0.8, 0.0, 0.0}

\definecolor{lightmauve}{rgb}{0.86, 0.82, 1.0}
\definecolor{lightpastelpurple}{rgb}{0.69, 0.61, 0.85}

\definecolor{ceil}{rgb}{0.57, 0.63, 0.81}
\definecolor{unitednationsblue}{rgb}{0.36, 0.57, 0.9}

\setul{0.5ex}{0.3ex}

\newcommand\Tstrut{\rule{0pt}{2.6ex}}         
\newcommand\Bstrut{\rule[-1.6ex]{0pt}{0pt}} 

\newcommand{\eatpunct}[1]{}

\DeclareRobustCommand*{\IEEEauthorrefmark}[1]{%
  \raisebox{0pt}[0pt][0pt]{\textsuperscript{\footnotesize #1}}%
}

\begin{document}

\title{\huge Background-Foreground Segmentation for Interior Sensing in Automotive Industry}

\author{\IEEEauthorblockN{Claudia Drygala\IEEEauthorrefmark{1},
Matthias Rottmann\IEEEauthorrefmark{1},
Hanno Gottschalk\IEEEauthorrefmark{1},\\
Klaus Friedrichs\IEEEauthorrefmark{2}
and
Thomas Kurbiel\IEEEauthorrefmark{2}
}
\\
\IEEEauthorblockN{\IEEEauthorrefmark{1}University of Wuppertal, School of Mathematics and Natural Sciences, IMACM \& IZMD}
\IEEEauthorblockN{\IEEEauthorrefmark{2}Aptiv Services Deutschland GmbH, Wuppertal, Germany}
\\
\IEEEauthorblockA{\tt \{drygala, rottmann, hgottsch\}@uni-wuppertal.de} 
\IEEEauthorblockA{\tt \{klaus.friedrichs, thomas.kurbiel\}@aptiv.com} 
}

\IEEEtitleabstractindextext{%
\begin{abstract}
To ensure safety in automated driving, the correct perception of the situation inside the car is as important as its environment. Thus, seat occupancy detection and classification of detected instances play an important role in interior sensing. By the knowledge of the seat occupancy status,  it is possible to, e.g., automate the airbag deployment control. Furthermore, the presence of a driver, which is necessary for partially automated driving cars at the automation levels two to four can be verified. In this work, we compare different statistical methods from the field of image segmentation to approach the problem of background-foreground segmentation in camera based interior sensing. 
In the recent years, several methods based on different techniques have been developed and applied to images or videos from different applications. The peculiarity of the given scenarios of interior sensing is, that the foreground instances and the background both contain static as well as dynamic elements. In data considered in this work, even the camera position is not completely fixed. 
We review and benchmark three different methods ranging, i.e., Gaussian Mixture Models (GMM), Morphological Snakes and a deep neural network, namely a Mask R-CNN. In particular, the limitations of the classical methods, GMM and Morphological Snakes, for interior sensing are shown. Furthermore, it turns, that it is possible to overcome these limitations by deep learning, e.g.\ using a Mask R-CNN. Although only a small amount of ground truth data was available for training, we enabled the Mask R-CNN to produce high quality background-foreground masks via transfer learning. Moreover, we demonstrate that certain augmentation as well as pre- and post-processing methods further enhance the performance of the investigated methods.
\end{abstract}

}

\maketitle

\thispagestyle{plain}
\pagestyle{plain}

\IEEEdisplaynontitleabstractindextext

\IEEEpeerreviewmaketitle

\section{Introduction}
\begin{figure}[tb]
    \centering
    \begin{minipage}{0.22\textwidth}
        \includegraphics[width=\textwidth]{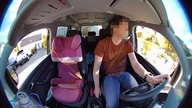}
    \end{minipage}
    \begin{minipage}{0.22\textwidth}
        \includegraphics[width=\textwidth]{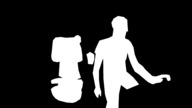}
    \end{minipage}\vspace{.1cm}
        \begin{minipage}{0.22\textwidth}
        \includegraphics[width=\textwidth]{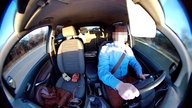}
    \end{minipage}
    \begin{minipage}{0.22\textwidth}
        \includegraphics[width=\textwidth]{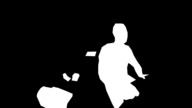}
    \end{minipage}
   \caption{Left: An RGB image of the test set. Right: The corresponding binary segmentation mask by which the foreground instances are separated of the background.}
   \label{fig:intro}
\end{figure}
Interior Sensing is of high importance  for automated driving. 
For instance, interior sensing aims at seat occupancy detection 
and classification \cite{interior_sensing_evaluation, dfki}. The classes may range from ``Person'', ``Child seat'' and ``Animal'' to ``Everyday object''.
This knowledge about the seat occupancy can be used e.g.\ for  smart airbag deployment control systems \cite{airbag}.
While the activation of the airbag could save a person's life in case of an accident, it could lead to serious injuries \cite{cs_airbag_injury_2003,cs_airbag_injury_2007} or even to death \cite{childseat_airbag_death_2005,cs_a_heat_death,american_RKI} for a child, which is sitting in a rear-facing child seat on the passenger seat. In the case of partially autonomous vehicles at the levels two to four (defined by \cite{SAE}), a driver has to be present in the car. Thus it is necessary to verify, if a person is present on the driver seat \cite{dfki}.  
Lastly, also the back seats of the car are of interest just as the front seats. Thinking of the so called ``Forgotten Baby Syndrome'' \cite{childseat, cs_a_heat_death} the system could give an alarm, if a forgotten child would be detected on a back seat.

In this work, we suggest seat occupancy detection by background-foreground segmentation methods. Only the extracted foreground instances, belonging to the classes ``Person'', ``Child Seat'' or ``Object'', should be considered for the classification. The motivation behind this approach is to realize the classification task independently of the car's interior features and thus achieve better generalization.

Background-foreground segmentation \cite{fgbgseg1, fgbgseg2, bgfgseg3}, also known as background-foreground detection \cite{bgfgdec1, bgfgdec2, bowman_overview}  or background subtraction \cite{bgsub1, bgsub2, bgsub3, gmm_z, bowman_overview}, is an intensively studied field in computer vision. In recent years, several methods have been developed addressing various scenarios. These methods are based on completely different techniques. As it is described extensively in the survey \cite{bowman_overview}, the approaches range from classical statistics based to modern methods, which incorporate deep convolutional neural networks \cite{goodfellow}.

The goal of this work is to introduce a dataset for the training of the background-foreground segmentation task and benchmark three methods in the given setting. The benchmark is performed on the quality of the generated background-foreground masks (see \cref{fig:intro}). Note that the minimization of the computational costs is not of interest for this work.
The three methods we selected are based on different techniques:
\begin{enumerate}
    \item Gaussian Mixture Model: A classical statistical method for background subtraction.
    \item Morphological Snakes: A classical approach to object detection bases on active contours.
    \item Mask R-CNN: A modern method solving the instance segmentation task by using deep neural networks.
\end{enumerate}
In this work, we investigate the limitations of each approach.
To ensure the comparability of the methods, all of them have been tested on a challenging test set of 100 real-world images. The test set is part of the dataset which is introduced by this work, named the ISSO dataset (Interior Sensing and Seat Occupany). The dataset consists of 1300 annotated real-world images extracted of videos recorded by employees of the company APTIV in Wuppertal, Germany which are  splitted into a training set of 1100 images, a validation and a test set of 100 images each. The images of the ISSO dataset describe scenarios of the interior of 13 different cars, as shown in \cref{fig:intro}. 
Scenarios of interior sensing are highly complex, since the foreground instances and the background can be both, dynamic and static. In this work, even the camera position varies slightly from car to car. Additionally, the impact of environmental effects has to be taken into account, like different weather conditions, shadows, traffic lights and vibrations.

Although only a rather small amount of 1100 real-world annotated images for the training is available, we demonstrate with the help transfer learning that it is possible to generate background-foreground masks of high quality with a Mask R-CNN.
Furthermore, we investigate to what extent the performance of the methods can be leveraged by certain pre- and post-processing methods for the data as well as by applying data augmentation techniques during the training of the neural network. In particular we study the effect of
\begin{itemize}
    \item the conversion to different color spaces (RGB, HSV, CIEL$^\ast$a$^\ast$b$^\ast$),
    \item contrast enhancement methods (Histogram Equalization, CLAHE),
    \item morphological operators (Closing, Opening) and
    \item data augmentation before and during the training of neural networks. 
\end{itemize}

The paper is organized as follows: The theoretical background to the methods considered is briefly explained in \cref{sec:methods}. This is followed by \cref{sec:data}, where the description of the data set designed, registered and annotated for this work is given. In \cref{sec:evaluation} the metrics by which the methods are evaluated are introduced. The results to the experiments are presented and discussed in \cref{sec:results}. In particular, the three methods are compared and the limitations of each method is discussed. Finally, we provide our conclusion and an outlook in \cref{sec:conclusion}.

\section{A Choice of Methods for Background-Foreground Segmentation}
\label{sec:methods}
\subsection{Gaussian Mixture Model (GMM)}
\label{ssec:GMM}
The GMM, introduced for foreground-background segmentation in \cite{gmm_c}, is based on the principle of background subtraction. As described in the \cref{ssec:color}, the values of a pixel are defined by a certain color space. For example, the pixel values of a gray scale image are given by single scalars, whereas the pixel values of a color image are given by a vector with the number of channels as dimension. In the GMM framework, these values of each pixel are modeled by a mixture of adaptive Gaussian distributions. The underlying data for the computation of those mixture models is given by a so called pixel process $\{X_1, \ldots, X_t\}$ which is a time series of pixel values. Thus, at any time $t$, the history for each pixel at position $(w_0,h_0)$ is known:
    \begin{equation}
        \{X_1, \ldots, X_t\} = \{I((w_0, h_0),i) | 1 \leq i \leq t\}
    \end{equation}
with $I$ being the image sequence. Now, the probability of observing a certain pixel value at time $t$ is defined as
    \begin{equation}
    p(X_t) = \sum \limits_{m=1}^K \hat{w}_{m,t} \mathcal{N}(X_t; \hat{\mu}_{m,t}, \hat{\Sigma}_{m,t})
    \end{equation}
    with
    \begin{itemize}
        \item $\mathcal{N}$: The multivariate Gaussian distribution.
        \item $K$: Number of Gaussian distributions in a mixture. 
        \item $\hat{\mu}_{m,t}$: The estimate of the mean value of the $m$-th Gaussian at time $t$.
        \item $\hat{\Sigma}_{m,t}$: The estimate of the covariance matrix of the $m$-th Gaussian at time $t$ defined as 
        $$
        \hat{\Sigma}_{m,t}=\hat{\sigma}^2_{m,t}\mathbf{I}
        $$
        with $\hat{\sigma}^2_{m,t}$  the estimate of the variance of the $m$-th Gaussian at time $t$ and $\mathbf{I}$ the identity matrix of appropriate dimension.
        \item $\hat{w}_{m,t}$: The estimated weight of the  $m$-th Gaussian at time $t$. Moreover, the weights fulfill the properties of non-negativity and normalization, i.e., $\hat{w}_{m,t} \geq 0$ and $\sum_{m=1}^M \hat{w}_{m,t} = 1$.
    \end{itemize}
If a new frame of the image sequence is considered at current time $t$, a new pixel value enters the pixel process. By the update of the pixel process, the estimates of the Gaussian distributions also have to be updated. For the estimation of the parameters, the Maximum Likelihood estimator for the currently observed data is computed. A well-known approach for this computation is the Expectation Maximization (EM) \cite{EM}. However, here the application of the exact EM algorithm would be costly, since the values of each pixel are modeled by a mixture of Gaussians and thus, the parameters are updated pixel-wise. Hence, the parameter update is realized by the implementation of an online $K$-means approximation \cite{gmm_c}. For each new pixel value it is checked whether the pixel is represented by one of the already existing $K$ Gaussians. This check is performed until a match is found. For example, a match is given if the new pixel value is within 2.5 standard deviations of a distribution.

\paragraph[Estimation of the model parameters]{\bf Estimation of the model parameters}
Whether there is a match or not, the weight parameters are iteratively updated via 
\begin{align}
    \hat{w}_{m,t} &=(1-\alpha)\hat{w}_{m,t-1}+\alpha\mathbb{M}_{m,t}\, .
    \label{eq:gmm_weights}
\end{align}
Here, $\alpha \in [0, 1]$ is the learning rate which determines the influence of data from past points in time and the speed at which the model parameters are updated and 
\begin{equation}
\mathbb{M}_{m,t}=
\begin{cases}
    1, ~~~{\rm in~case~of~a~match}\\
    0, ~~~{\rm else.}
\end{cases}
\end{equation}
The distribution parameters $\hat{\mu}$ and $\hat{\sigma}^2$ are only updated for the distribution that matches the new pixel value $X_t$, otherwise no update is performed:
\begin{equation}
    \hat{\mu}_{m,t}= 
    \begin{cases}
    (1-\rho)\hat{\mu}_{m,t-1} +\rho X_t, ~~~\mathbb{M}_{m,t}=1\\
    \hat{\mu}_{m,t-1},  ~~~~~~~~~~~~~~~~~~~~{\rm else}
    \end{cases}
\end{equation}

\begin{equation}
    \hat{\sigma}^2_{m,t}=
    \begin{cases}
     (1-\rho)\hat{\sigma}^2_{m,t-1}+\rho \hat{\delta}_{m,t}^T\hat{\delta}_{m,t}, ~\mathbb{M}_{m,t}=1\\
     \hat{\sigma}^2_{m,t-1}, ~~~~~~~~~~~~~~~~~~~~~~~~~{\rm else}
    \end{cases}
\end{equation}
with $\rho=\alpha\mathcal{N}(X_t|\hat{\mu}_{m,t}, \hat{\sigma}^2_{m,t})$ and $\hat{\delta}_{m,t}=X_t-\hat{\mu}_{m,t}$.
If no match is given at all, the distribution that assigns the lowest probability to the data is replaced by a new distribution with the initial parameters $\hat{w}_{new}=\alpha$,  $\hat{\mu}_{new}=X_t$, $\hat{\sigma}_{new}=\sigma_0$ and $\sigma_0$ an appropriate initial variance \cite{gmm_z, gmm_c}.

\paragraph[Estimation of the background model]{\bf Estimation of the background model}
Now, it should be determined by which of the computed Gaussian distributions, the background can be modeled.  In particular, the Gaussians with the highest weights and the lowest variances are of interest. Generally, it can be assumed that pixel values describing the background of a scenario are repeated and thus, also their distributions. Hence, if a new pixel value enters to a pixel process which describes the background, a high probability for a match is given. By the update rule \eqref{eq:gmm_weights}, it can be observed that the weights are increasing in the case of a match. Moreover, the background consists mostly of static elements that produce less variance than dynamic ones.
Therefore, to determine the distributions of the mixture model that describe the background the best, the Gaussians are sorted firstly in descending order by the value $\nicefrac{\hat{w}}{\hat{\sigma}}$. Hence, the distributions which are most likely representing the background are  at the top of the list. Then, the first $B$ distributions are chosen to model the background
\begin{equation}
    B=\underset{b}{\rm argmin} \left( \sum\limits_{m=1}^b \hat{w}_m > \tau \right)
\end{equation}
with $\tau$ the percentage of the pixel process that should affect the background model.
The pixel values that cannot be assigned to a distribution which belongs to the background model are grouped by a two-pass connected components algorithm \cite{concomp}.

\paragraph[Number of mixtures]{\bf Number of mixtures}
In the introduced GMM framework of above, the number of Gaussian distributions in a mixture is given by a constant value  that is determined by the available memory and computational power. In this work, a modified version of the original GMM framework is used, where the number of Gaussians is also adaptive. In \cite{gmm_z} the update rule of $\hat{w}$ is reformulated such that the weights may take negative values. This aims omitting weights for Gaussians which are not relevant for the background estimation. Hence, the distributions that do not describe the background with high certainty are directly excluded. We refer to \cite{gmm_z} for a detailed derivation of the modified update rule. 

\subsection{Morphological Snakes}\label{sec:morph_snakes}
Originally, the object detection method based on active contours (also called ``snakes'') was presented in \cite{snakes_kass}. The idea behind this approach is to detect foreground instances of an image $I$ by evolving an initial curve $C_0$ towards the instances boundaries. In particular, the evolution of this curve is achieved by minimizing the energy functional 
\begin{align}
\begin{split}
E(C) &= \alpha \int_0^1 \lVert C'(q) \rVert_2 ^2 dq + \beta \int_0^1 \lVert C''(q) \rVert_2 ^2 dq\\
&- \lambda \int_0^1 \lVert \nabla I(C(q))  \rVert_2 dq
\label{eq:snakes}
\end{split}
\end{align}
with $C(q):[0,1]\rightarrow  \mathbb{R}^2$ a parameterized planar curve, which represents the contour, $I:[0,a]\times [0,b] \rightarrow \mathbb{R}^+$ the considered image, $a, b \in \mathbb{R}^+$ and $\alpha, \beta, \lambda \in \mathbb{R}^+$ constant parameters. 
By the design of the functional, the smoothness of the curve is controlled by the first two terms, while the third term attracts the curve towards the boundary of the object. Therein, the gradient of the image $\nabla I$ acts as an edge detector. Hence, the (local) minimum should be obtained at the objects boundary.

To handle topological changes, such as splitting and merging, automatically, the original energy functional is modified in different ways. The ``Geodesic Active Contours'' (GAC) \cite{GAC} and the ``Active Contours Without Edges'' (ACWE) \cite{ACWE} are based on the level-set-method \cite{Osher} which are successfully applied to conduct curve evolution \cite{GAC}. By this, the curve $C: [0,1] \times \R^+ \rightarrow \mathbb{R}^2,~(q,t)\mapsto C(q,t)$ parameterized over time $t\in \mathbb{R}^+$ is included into a level-set of an arbitrary smooth embedding function $u:\R^2 \times \R^+ \rightarrow \R$, such that it holds $C(q,t)=\{(x,y) | u((x,y);t)=0\}$. Hence, the curve $C$ is represented implicitly by $u$ \cite{morphsnakes_detail}.

To receive the level-set formulation, the evolution of $C$ is defined by a partial differential equation (PDE) $C_t$ obtained by minimizing the respective energy functional $E(C)$ with the steepest descent method. Then, this curve evolution $C_t$ can be reformulated into the level-set equation $u_t=\frac{\partial u}{\partial t}$ for $t>0$ with the initial value $u_0=u((x,y);0)$, as shown in \cite{GAC, numOS}.
For the GAC- and ACWE method, this approach results in the following level-set equations for $t>0$.

\paragraph[Geodesic Active Contours (GAC)]{\bf Geodesic Active Contours (GAC)}
For this approach, the level-set equation is given by
\begin{equation}
    \label{eq:gacLevelSet}
    u_t = g(I) \tilde{\kappa} \lVert \nabla u \rVert_2 + g(I)v \lVert \nabla u \rVert_2  + \nabla g(I) \nabla u
\end{equation}
with
\begin{itemize}
    \item $I:[0,a]\times [0,b] \rightarrow \R^+, a,b \in \R^+$ an image.
    \item $g(I):[0,\infty) \rightarrow \R^+$ a strictly decreasing function. By the values of $g$, the image regions of interest can be selected, e.g.\ the object boundaries in the case of image segmentation. In this work, $g$ is defined as
    \begin{equation}
        g(I)=\frac{1}{\sqrt{1+\alpha\lVert G_\sigma \ast I \rVert}}
        \label{eq:mgac_filter}
    \end{equation}
    with $G_\sigma \ast I $ a Gaussian filter ($\ast$ being the convolution operator), $\sigma$ the standard
    deviation and $\alpha>0$ a non-linear scaling parameter. On object boundaries, $g(I)$ takes smaller values than on homogeneous image areas.
    \item $\tilde{\kappa} := {\rm div} \left( \frac{\nabla u}{\lVert \nabla u \rVert} \right)$ the euclidean curvature of the embedding function $u$, proven in \cite{numOS}.
    \item $v \in \R$ the balloon force parameter.
\end{itemize}
\paragraph[Active Contours Without Edges (ACWE)]{\bf Active Contours Without Edges (ACWE)}
Herein, the level-set equation is given by
\begin{equation}
    \label{eq:acweLevelSet}
    u_t= \lVert \nabla u \rVert_2   \left(\mu \tilde{\kappa} -v- \lambda_1(I-c_1)^2 +\lambda_2(I-c_2)^2\right)
\end{equation}
with 
\begin{itemize}
    \item $I$, $\tilde{\kappa}$ and $v$ analogous to above.
    \item $c_1, c_2$ constants that depend on the curve $C: [0,1] \rightarrow \R^2$. In particular, $c_1$ represents the average of the pixel values of $I$ inside $C$ and  $c_2$ the average of $I(x,y)$ outside $C$.
    \item $\lambda_1, \lambda_2 >0$ and $\mu \geq 0$ fixed weight parameters.
\end{itemize}
Both level-set equations are composed of a smoothing term, a balloon force term and an attraction force or image attachement term. In particular, the parts of the curve with a high curvature will be smoothed by the smoothing term. The balloon force term should help to accelerate the curve evolution especially in areas, where the attraction force is too weak due to small values of the gradients in less informative areas. So, the evolution of $C$ is inflated ($v>0$) or deflated ($v<0$) by determining the velocity $v \in \R$. If $v=0$, the balloon force is switched off.

Now, the solutions of the level-set equations \eqref{eq:gacLevelSet} and \eqref{eq:acweLevelSet}  are obtained by solving the time-dependent PDEs iteratively. However, the numerical methods for the computation of PDEs are costly, challenging in the implementation and suffer from stability constraints. In \cite{morphsnakes_detail} it is shown, that it is possible to overcome these difficulties by approximating the PDEs of which \eqref{eq:gacLevelSet} and \eqref{eq:acweLevelSet} are composed of, by binary morphological operators. These operators are formulated as $\mathrm{sup}$-$\mathrm{inf}$ operators as given in \eqref{eq:supInf}. The implementation of such a $\mathrm{sup}$-$\mathrm{inf}$ operator is much easier and the computation is more stable and faster compared to the one of PDEs. Hence, the PDEs \eqref{eq:gacLevelSet} and \eqref{eq:acweLevelSet} are approximated by the composition of mathematical morphological operators, whereby the implicit representation of $C$ is maintained.

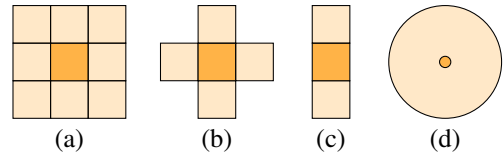
\begin{figure}[!h]
    \centering
    \begin{tikzpicture}
    \node[draw, rectangle, minimum width=0.5cm, minimum height=0.5cm, fill=porange, fill opacity=0.3] at (0,0) {};
    \node[draw, rectangle, minimum width=0.5cm, minimum height=0.5cm, fill=porange, fill opacity=0.3] at (0,0.5) {};
    \node[draw, rectangle, minimum width=0.5cm, minimum height=0.5cm, fill=porange, fill opacity=0.3] at (0,1) {};
    \node[draw, rectangle, minimum width=0.5cm, minimum height=0.5cm, fill=porange, fill opacity=0.3, label={[label distance=0.01cm]270:(a)}] at (0.5,0) {};
    \node[draw, rectangle, minimum width=0.5cm, minimum height=0.5cm, fill=porange] at (0.5,0.5) {};
    \node[draw,rectangle, minimum width=0.5cm, minimum height=0.5cm, fill=porange, fill opacity=0.3] at (0.5,1) {};
    \node[draw,rectangle, minimum width=0.5cm, minimum height=0.5cm, fill=porange, fill opacity=0.3] at (1,0) {};
    \node[draw,rectangle, minimum width=0.5cm, minimum height=0.5cm, fill=porange, fill opacity=0.3] at (1,0.5) {};
    \node[draw, rectangle, minimum width=0.5cm, minimum height=0.5cm, fill=porange, fill opacity=0.3] at (1,1) {};
    \end{tikzpicture}
    \hspace{2mm}
    \begin{tikzpicture}
    \node[draw, rectangle, minimum width=0.5cm, minimum height=0.5cm, fill=porange, fill opacity=0.3] at (0,0.5) {};
    \node[draw, rectangle, minimum width=0.5cm, minimum height=0.5cm, fill=porange, fill opacity=0.3, label={[label distance=0.01cm]270:(b)}] at (0.5,0) {};
    \node[draw, rectangle, minimum width=0.5cm, minimum height=0.5cm, fill=porange] at (0.5,0.5) {};
    \node[draw,rectangle, minimum width=0.5cm, minimum height=0.5cm, fill=porange, fill opacity=0.3] at (0.5,1) {};
    \node[draw,rectangle, minimum width=0.5cm, minimum height=0.5cm, fill=porange, fill opacity=0.3] at (1,0.5) {};
    \end{tikzpicture}
    \hspace{2mm}
    \begin{tikzpicture}
    \node[draw, rectangle, minimum width=0.5cm, minimum height=0.5cm, fill=porange, fill opacity=0.3, label={[label distance=0.01cm]270:(c)}] at (0.5,0) {};
    \node[draw, rectangle, minimum width=0.5cm, minimum height=0.5cm, fill=porange] at (0.5,0.5) {};
    \node[draw,rectangle, minimum width=0.5cm, minimum height=0.5cm, fill=porange, fill opacity=0.3] at (0.5,1) {};
    \end{tikzpicture}
    \hspace{2mm}
    \begin{tikzpicture}
    \node[draw, circle, minimum size=1.5cm, fill=porange, fill opacity=0.3, label={[label distance=0.01cm]270:(d)}] at (0,0) {};
    \node[draw, circle, scale=0.45, fill=porange] at (0,0) {};
    \end{tikzpicture}
    \caption{Examples for the shape of a structuring element $B$: (a) Square, (b) diamond, (c) line segment and (d) ball. By the darker shaded areas, the origin of $B$ is described. According to \cite{ErosionDilation}.}
    \label{fig:example_struct_elem}
\end{figure}

Generally, a morphological operator $T$ satisfies the properties of standard monotony, translation- and contrast invariance \cite{GCE}. Furthermore, $T$ is defined uniquely by a structuring element $B$, which is a set of arbitrary but small size and shape with a predefined origin as described in \cref{fig:example_struct_elem}. Usually, $B$  is significantly smaller than the considered image $I$. Mathematically, $B$ is a matrix of dimension $c \times d$, $c,d \geq 1$ consisting of zeros and ones. The  role of $B$ is to probe a given image pixel-wise whereby the positioning of $B$ at a pixel is given by its defined origin. According to the rule which is specified by a morphological operator, every pixel is evaluated by the comparison with the origin of $B$ and its corresponding neighborhood which is represented by the values  equal to one in the matrix \cite{ErosionDilation}. Hence, the structuring element can be interpreted as a kernel in context of machine learning \cite{GCE}.

Morphological operators of interest for this work are the dilation, the erosion and the curvature morphological operators, due to their properties regarding the infinitesimal behaviour.

\begin{remark}{{\rm \bf (Notation)}}\\
Let $\mathcal{F} \subseteq C_b^k(\R^n)$ be a set of bounded continuous differentiable functions up to order $k$ over $\R^n$. The function operator is denoted by $T:\mathcal{F} \rightarrow \mathcal{F}$ which is assumed to be well-defined on $ C_b^k(\R^n)$ \cite{axioms}.
\end{remark}
\begin{definition}{{\rm {\bf (Dilation and Erosion} \cite{morphTh}{\bf)}}}\\
Let $u \in \F$, $B$ the structuring element and $h\geq 1, h\in \R$ a scaling parameter. The \underline{Dilation} of $u$ by $hB$, written as $D_{h}=D_{hB}$, is defined by
\begin{equation}
D_{h}u(\x) = \sup_{\y \in hB} u(\x-\y). 
\end{equation}
Whereas, the \underline{Erosion} of $u$ by $hB$, written as $E_{h}=E_{hB}$, is given by
\begin{equation}
E_{h}u(\x) = \inf_{\y \in -hB} u(\x-\y). 
\end{equation}
\end{definition}
\vspace{-5mm}
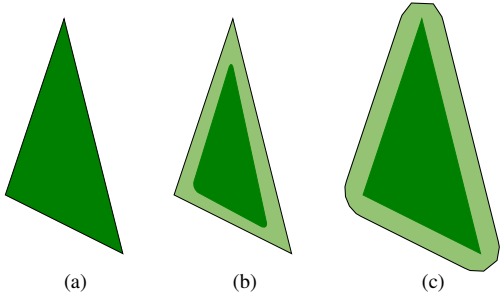
\begin{figure}[!h]
    \centering
    \resizebox{!}{4cm}{%
    \begin{tikzpicture}
        \path (-1,-1) coordinate (A) (1,-2) coordinate (B) (0,2) coordinate (C);
        \draw[fill=ao] (A) -- (B) -- (C) -- cycle;
        \node[text width=1cm] at (0.5, -2.5) {(a)};
    \end{tikzpicture}
    \hspace{5mm}
        \begin{tikzpicture}
      \path (-1,-1) coordinate (A) (1,-2) coordinate (B) (0,2) coordinate (C);
      \path (-0.7,-0.9) coordinate (A1) (0.6,-1.6) coordinate (B2) (-0.025,1.3) coordinate (C3);
      \draw[fill=pistachio] (A) -- (B) -- (C) -- cycle;
      \draw[ao, rounded corners, fill=ao] (A1) -- (B2) -- (C3) -- cycle;
      \node[text width=1cm] at (0.5, -2.5) {(b)};
    \end{tikzpicture}
    \hspace{5mm}
       \begin{tikzpicture}
      \path (-1,-1) coordinate (A) (1,-2) coordinate (B) (0,2) coordinate (C);
      \draw[rounded corners, line width=20pt] (A) -- (B) -- (C) -- cycle;
      \draw[rounded corners, pistachio, line width=19pt] (A) -- (B) -- (C) -- cycle;
      \draw[ao, fill=ao] (A) -- (B) -- (C) -- cycle;
      \node[text width=1cm] at (0.5, -2.5) {(c)};
    \end{tikzpicture}
 }
    \caption{The effect of Erosion (b) and Dilation (c) applied on an object (a) with a ball as structuring  element $B$. According to \cite{ErosionDilation, GCE}.}
    \label{fig:DilationErosion}
\end{figure}

\paragraph[Infinitesimal behaviour of Dilation and Erosion]{\bf Infinitesimal behaviour of Dilation and Erosion}
Let be $B$ convex bounded and the $B$-norm on $\R^n$ is defined by 
$\lVert \x \rVert_B = \sup_{\y \in B} (\x \cdot \y)$ with $\cdot$ the Euclidean scalar product. Furthermore, the initial value of $u$ at time $t=0$ is given by $u_0(\x)=u(\x;0)$ with $\x  \in \R^n$.
By defining $u:\R^n \times \R^+ \rightarrow \R$ as the dilation of the initial value $u_0$ by $tB$, such that $u(\x;t)=D_t u_0(\x)$, it holds (see \cite{morphTh, axioms}):
\begin{equation}
\dfrac{\partial u}{\partial t} =\lVert \nabla u \rVert_B.
\label{eq:dilationPDE}
\end{equation}
Analogously it holds for $u(\x;t)=E_t u_0(\x)$ (see \cite{morphTh, axioms}):
\begin{equation}
\dfrac{\partial u}{\partial t} =-\lVert \nabla u \rVert_B.
\label{eq:erosionPDE}
\end{equation}
Here, it holds in particular that 
\begin{equation}
    \lVert \nabla u \rVert_B=\lVert \nabla u \rVert_2
\end{equation}
if the structuring element is defined as the unit ball 
\begin{equation}
    B_1(0)=\{\x\in \R^n: \lVert \x \rVert_2 <1\}
\end{equation}
on $\R^n$ \cite{norms, morphTh}.
Hence, under certain conditions, the infinitesimal behaviour of the Dilation and the Erosion is equivalent to the PDE $\frac{\partial u}{\partial t}=\pm \lVert \nabla u \rVert_2$, which is a component of the level-set equations \eqref{eq:gacLevelSet} and \eqref{eq:acweLevelSet}.

\paragraph[The Sup-Inf Representation of Morphological Operators]{\bf The Sup-Inf Representation of Morphological Operators}
The authors of \cite{morphTh} show that every morphological operator has a $\mathrm{sup}$-$\mathrm{inf}$ representation and that also the dual $\mathrm{inf}$-$\mathrm{sup}$ form exists.

Let $\mathcal{B}$ be a set of structuring elements and $T:\mathcal{F}\rightarrow\mathcal{F}$ an arbitrary morphological operator. Then $T$ can be represented  by the sup-inf operator
\begin{equation}
SI_h := T_h u(\x)=\sup_{B\in \mathcal{B}} \inf_{\y \in \x+hB} u(\y).
\label{eq:supInf}
\end{equation}
The dual operator of $T$ is defined as $\tilde{T}(u)=-T(-u)$, which is also a morphological operator. Thus, the inf-sup representation of $T$ is given by 
\begin{equation}
IS_h:= T_h u(\x)=\inf_{B\in \mathcal{\tilde{B}}} \sup_{\y \in \x+hB} u(\y)
\label{eq:infSup}
\end{equation}
with $\tilde{\mathcal{B}}$ the set of structuring elements of $\tilde{T}$.
\begin{definition}{\rm {\bf (Curvature morphological operator} \cite{morphsnakes_detail}{\bf)}}\\
\label{def:curvatureMorphologicalOperator}
Given the morphological operators $SI_h$ and $IS_h$ with the set of structuring elements $\mathcal{B}=\{[-1,1]_\theta \subset \R^2; \theta\in [0,\pi) \}$ and $h$ sufficiently small. Then the composition 
\begin{equation}
    SI_{\sqrt{h}} \circ IS_{\sqrt{h}}
    \label{eq:curvMorphOper}
\end{equation}
is defined as the curvature morphological operator.
\end{definition}

\paragraph[Infinitesimal behaviour of $SI_{\sqrt{h}} \circ IS_{\sqrt{h}}$]{\bf Infinitesimal behaviour of $\mathbf{SI_{\sqrt{h}} \circ IS_{\sqrt{h}}}$}
It is shown in \cite{morphTh}, that the mean operator
\begin{equation}
F_h u(\x) = \dfrac{SI_{2h}(\x)+IS_{2h}u(\x)}{2}
\label{eq:meanOper}
\end{equation}
has an infinitesimal behaviour, which is equivalent to the mean curvature motion $\tilde{\kappa}\lVert \nabla u \rVert_2$. However, the problem about $F_h$ or rather $F_{\sqrt{h}}$ is, that it is not a morphological operator since the property of contrast invariance is not satisfied \cite{CatteDibosKoepfler}. Due to this reason, the curvature morphological operator is introduced by \cite{morphsnakes_detail}, which approximates the mean operator.
Thus, \eqref{eq:curvMorphOper} has the same infinitesimal behaviour as \eqref{eq:meanOper}, namely  $\tilde{\kappa}\lVert \nabla u \rVert_2$, which is also a component of the level-set equations \eqref{eq:gacLevelSet} and \eqref{eq:acweLevelSet}.

\paragraph[Morphological GAC (MGAC) and ACWE (MACWE)]{\bf Morphological GAC (MGAC) and ACWE (MACWE)}
Summarizingly, the introduced morphological operators approximate PDEs, which are defined by the level-set equations $u_t$ of the GAC and ACWE method. By this knowledge, it is possible to derive the morphological versions of the both methods \cite{morphsnakes_detail}.

\begin{figure}[!h]
    \centering
    \begin{tikzpicture}
    \node[draw, rectangle, minimum width=0.5cm, minimum height=0.5cm, fill=cblue] at (0,0) {};
    \node[draw, rectangle, minimum width=0.5cm, minimum height=0.5cm] at (0,0.5) {};
    \node[draw, rectangle, minimum width=0.5cm, minimum height=0.5cm] at (0,1) {};
    \node[draw, rectangle, minimum width=0.5cm, minimum height=0.5cm] at (0.5,0) {};
    \node[draw, rectangle, minimum width=0.5cm, minimum height=0.5cm, fill=navy] at (0.5,0.5) {};
    \node[draw, rectangle, minimum width=0.5cm, minimum height=0.5cm] at (0.5,1) {};
    \node[draw, rectangle, minimum width=0.5cm, minimum height=0.5cm] at (1,0) {};
    \node[draw, rectangle, minimum width=0.5cm, minimum height=0.5cm] at (1,0.5) {};
    \node[draw, rectangle, minimum width=0.5cm, minimum height=0.5cm, fill=cblue] at (1,1) {};
    \end{tikzpicture}
    \hspace{2mm}
    \begin{tikzpicture}
    \node[draw, rectangle, minimum width=0.5cm, minimum height=0.5cm] at (0,0) {};
    \node[draw, rectangle, minimum width=0.5cm, minimum height=0.5cm] at (0,0.5) {};
    \node[draw, rectangle, minimum width=0.5cm, minimum height=0.5cm] at (0,1) {};
    \node[draw, rectangle, minimum width=0.5cm, minimum height=0.5cm, fill=cblue] at (0.5,0) {};
    \node[draw, rectangle, minimum width=0.5cm, minimum height=0.5cm, fill=navy] at (0.5,0.5) {};
    \node[draw, rectangle, minimum width=0.5cm, minimum height=0.5cm, fill=cblue] at (0.5,1) {};
    \node[draw, rectangle, minimum width=0.5cm, minimum height=0.5cm] at (1,0) {};
    \node[draw, rectangle, minimum width=0.5cm, minimum height=0.5cm] at (1,0.5) {};
    \node[draw, rectangle, minimum width=0.5cm, minimum height=0.5cm] at (1,1) {};
    \end{tikzpicture}
    \hspace{2mm}
    \begin{tikzpicture}
    \node[draw, rectangle, minimum width=0.5cm, minimum height=0.5cm] at (0,0) {};
    \node[draw, rectangle, minimum width=0.5cm, minimum height=0.5cm] at (0,0.5) {};
    \node[draw, rectangle, minimum width=0.5cm, minimum height=0.5cm, fill=cblue] at (0,1) {};
    \node[draw, rectangle, minimum width=0.5cm, minimum height=0.5cm] at (0.5,0) {};
    \node[draw, rectangle, minimum width=0.5cm, minimum height=0.5cm, fill=navy] at (0.5,0.5) {};
    \node[draw, rectangle, minimum width=0.5cm, minimum height=0.5cm] at (0.5,1) {};
    \node[draw, rectangle, minimum width=0.5cm, minimum height=0.5cm, fill=cblue] at (1,0) {};
    \node[draw, rectangle, minimum width=0.5cm, minimum height=0.5cm] at (1,0.5) {};
    \node[draw, rectangle, minimum width=0.5cm, minimum height=0.5cm] at (1,1) {};
    \end{tikzpicture}
    \hspace{2mm}
    \begin{tikzpicture}
    \node[draw, rectangle, minimum width=0.5cm, minimum height=0.5cm] at (0,0) {};
    \node[draw, rectangle, minimum width=0.5cm, minimum height=0.5cm, fill=cblue] at (0,0.5) {};
    \node[draw, rectangle, minimum width=0.5cm, minimum height=0.5cm] at (0,1) {};
    \node[draw, rectangle, minimum width=0.5cm, minimum height=0.5cm] at (0.5,0) {};
    \node[draw, rectangle, minimum width=0.5cm, minimum height=0.5cm, fill=navy] at (0.5,0.5) {};
    \node[draw, rectangle, minimum width=0.5cm, minimum height=0.5cm] at (0.5,1) {};
    \node[draw, rectangle, minimum width=0.5cm, minimum height=0.5cm] at (1,0) {};
    \node[draw, rectangle, minimum width=0.5cm, minimum height=0.5cm, fill=cblue] at (1,0.5) {};
    \node[draw, rectangle, minimum width=0.5cm, minimum height=0.5cm] at (1,1) {};
    \end{tikzpicture}
    \caption{A discrete set of structuring elements $\mathcal{B}_d$ with the origin at the center. Adapted from: \cite{morphsnakes_detail}.}
    \label{fig:discrete_structuring_element}
\end{figure}
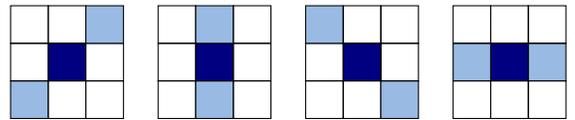

To this end, the embedding function $u$ needs to be redefined. Firstly, $u$ should be discrete in practice and secondly, $u$ has to be binary, since the morphological operators are also binary. Hence, $u:\mathbb{Z}^2 \rightarrow\{0,1\}$ is defined as a binary piece-wise constant function with 
\begin{equation}
u(\x)=
\begin{cases}
1,\quad {\rm if}~\x~{\rm~ is ~inside~the~curve~boundaries,}\\
0,\quad {\rm if}~\x~{\rm~ is~outside~the~curve~boundaries.}
\end{cases}
\label{eq:binaryDiscreteCurve}
\end{equation}
Due to the discretization of $u$, the (sets of) structuring elements also have to be discretized. This realizes the discretization of the morphological operators.
In \cref{fig:discrete_structuring_element}, a possible discrete version $\mathcal{B}^d$ of $\mathcal{B}$ in definition \ref{def:curvatureMorphologicalOperator} is described. Here, $\mathcal{B}^d$ consists of four discrete line segments with a length of three pixels and the origin at the pixel coordinate (0,0). Analogously, the structuring element of the Dilation and the Erosion given by the unit ball $B_1(0)$ can be discretized.

Intuitively, the balloon force operator acts in a similar way as the Dilation or the Erosion by inflating or deflating a contour, respectively (see \cref{fig:DilationErosion}). So the PDE of the balloon force term can be approximated by the Dilation, if $v>0$ and vice versa by the Erosion.
The smoothing term represents the mean curvature motion, such that the PDE of this component is approximated by the curvature morphological operator.
Finally, the remaining attraction force $\nabla g(I) \nabla u$ and image attachment term $\lVert \nabla u \rVert_2   \left(\lambda_2(I-c_2)^2-\lambda_1(I-c_1)^2\right)$ can be discretized directly, as the remaining factors $g(I)$ of the balloon and the smoothing term. In \cite{morphsnakes_detail} it is described how this discretization is realized.

In conclusion, the level-set equations \eqref{eq:gacLevelSet} and \eqref{eq:acweLevelSet} are solved by the successive computation of the composition of three discrete and morphological operators in the mGAC or the mACWE approach. The algorithms of both approaches can be found in \cite{morphsnakes_detail}.

\subsection{Mask R-CNN}
\tikzset{halo/.style={rounded rectangle, minimum width=6.5cm, minimum height=1.5cm, fill=orange, opacity=0.15}}
\tikzset{halo_text/.style={rounded rectangle, minimum width=2cm, minimum height=0.5cm, fill=orange, opacity=0.15}}
\begin{figure*}[ht]
    \centering
\begin{tikzpicture}
\node[inner sep=0pt,  minimum width=1cm, minimum height=1cm, label={below:\small Input image}] (i) at (0,0){\includegraphics[width=.12\textwidth]{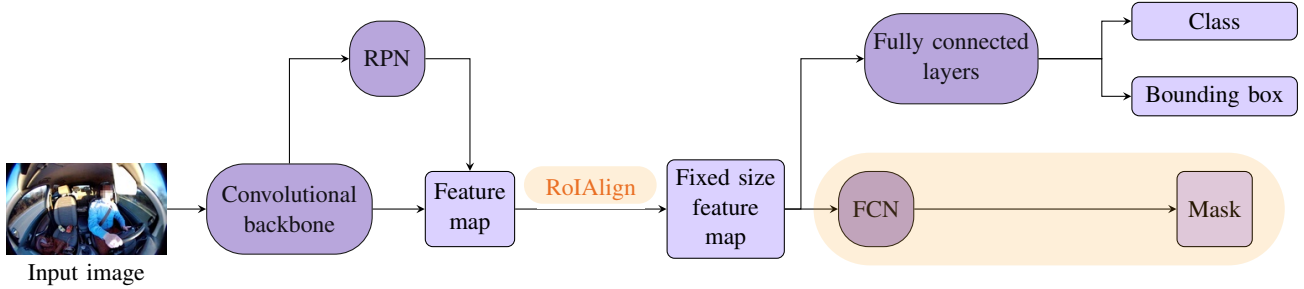}};
\node[draw, rectangle, rounded corners=0.5cm, fill=lightpastelpurple, opacity=0.9, align=center, minimum width=2.2cm, minimum height=1.2cm] (cb) at (2.7,0) {\small Convolutional\\ \small backbone};
\draw[-stealth](i)--(cb);
\node[draw,rectangle,rounded corners=0.1cm , fill=lightmauve, align=center, minimum width=1cm, minimum height=1cm] (fm) at (5.1,0)  {\small Feature\\ \small map};
\draw[-stealth](cb)--(fm);
\node[draw, rectangle, rounded corners=0.4cm, fill=lightpastelpurple, opacity=0.9, align=center, minimum width=1cm, minimum height=1cm] (rpn) at (4,2)  {\small RPN};
\draw[-stealth](cb.north)-- (2.7,2) --  (rpn.west);
\draw[-stealth](rpn.east)-- (5.1,2) --  (fm.north);
\node[draw,rectangle,rounded corners=0.1cm , fill=lightmauve, align=center, minimum width=1cm, minimum height=1cm] (fsfm) at (8.5,0)  {\small Fixed size\\ \small feature\\ \small map};
\draw[-stealth](fm) -- (fsfm)  node [midway, above, fill=white] {\small \textcolor{deepcarrotorange}{RoIAlign}};
\node[draw, rectangle, rounded corners=0.4cm, fill=lightpastelpurple, opacity=0.9, align=center, minimum width=1cm, minimum height=1cm] (fcn) at (10.5,0)  {\small FCN};
\draw[-stealth](fsfm) --  (fcn);
\node[draw,rectangle, rounded corners=0.1cm , fill=lightmauve, align=center, minimum width=1cm, minimum height=1cm] (m) at (15,0)  {\small Mask};
\draw[-stealth](fcn) -- (m);
\node[draw,rectangle, rounded corners=0.5cm, fill=lightpastelpurple, opacity=0.9, align=center, minimum width=2.2cm, minimum height=1.2cm] (fc) at (11.5,2)  {\small Fully connected\\ \small layers};
\draw[-stealth](fsfm.east) -- (9.5,0) -- (9.5,2) -- (fc);
\node[draw,rectangle, rounded corners=0.1cm , fill=lightmauve, align=center, minimum width=2.2cm, minimum height=0.5cm] (bb) at (15,1.5)  {\small Bounding box};
\node[draw,rectangle, rounded corners=0.1cm , fill=lightmauve, align=center, minimum width=2.2cm, minimum height=0.5cm] (reg) at (15,2.5)  {\small Class};
\draw[-stealth](fc.east) -- (13.5,2) -- (13.5,1.5) -- (bb);
\draw[-stealth](fc.east) -- (13.5,2) -- (13.5,2.5) -- (reg);

\node[halo] at (12.8,0) {};
\node[halo_text] at (6.7,0.32) {};
\end{tikzpicture}
    \caption{Architecture of the Mask R-CNN. The orange shaded parts represent the extensions of the Faster R-CNN by which the Mask R-CNN is derived (according to \cite{mrcnn, faster_rcnn}).}
    \label{fig:MaskRCNN}
\end{figure*}

The Mask R-CNN \cite{mrcnn} is a Region-based Convolutional Neural Network for instance segmentation. Hence, the goal is to detect and classify each object of an image, whereby it should be also distinguished between every individual instance within a class. By a Mask R-CNN, the detection and classification task and the generation of a mask for each instance are managed simultaneously.

In particular, the Mask R-CNN is an extension of the Faster R-CNN \cite{faster_rcnn}. This framework for object detection consists of two components: A Region Proposal Network (RPN) and a region-based object detection network, here given by the Fast R-CNN \cite{fast_rcnn} (the predecessor of Faster R-CNN). The RPN generates candidate object locations, called ``proposals'', which the Fast R-CNN uses to determine the exact locations of the detected images. The innovation of the Faster R-CNN is to unify those both networks into one framework by developing training algorithms in which both networks share some of their layers. Now, the extension of the Faster R-CNN is realized by adding a mask branch. As described in \cref{fig:MaskRCNN}, the instance masks are generated by a Fully Convolutional Network (FCN) \cite{fcn} which aims at classifying
for each pixel, whether it belongs to a certain class or not.

For the achievement of the prediction of high quality masks, the authors of \cite{mrcnn} show, that the introduction of the following two novelties play a key role.
Firstly, the ``RoIPooling''-layer  (RoI: Region of Interest) of the Faster R-CNN is substituted by the ``RoIAlign''-layer. Actually, the Faster R-CNN is not designed for a pixel-to-pixel relation between the in- and output. By RoIAlign the features which are extracted by the convolutional backbone, can be properly aligned according to the input image (see \cref{fig:MaskRCNN}). Thus, the generation of pixel-accurate instance segmentation masks is possible.
Secondly, the classification task and the prediction of the mask for each instance is decoupled. The loss function is defined in such a manner, that binary masks are predicted for all of the $K$ classes independently, such that no competition exists among these classes during inference. Hence, the prediction of the class is not based on a predicted mask, but solely on the classification branch. Since all desired outputs are computed in parallel, the multi-task loss 
\begin{equation}
    \L = \L_C + \L_B + \L_M
\end{equation}
is defined on each RoI which is composed of the classification loss $\L_C$ \cite{fast_rcnn}, the bounding box regression loss $\L_B$ \cite{fast_rcnn} and the loss of the mask branch $\L_M$ \cite{mrcnn,crossentropy}. These loss functions are defined as described below.
\begin{remark}{\rm \bf (Notation)}\\
The set of ground truth labels is given by $\mathcal{Y}$, whereby $\# \mathcal{Y} = K+1$, and consists of $K$ predefined object classes and an additional background class. In particular, the background class is denoted by $y=0$. Moreover, the ground truth class $y$ of each instance is assigned to each RoI.
\end{remark}
\paragraph[1. Classification loss $\L_C$]{\it 1. Classification loss $\L_C$}
The output of the classification branch is given by a discrete probability distribution $p = (p_0, p_1, \ldots, p_y \ldots, p_K)$ over all $K+1$ classes whereby 
\begin{equation}
    p_k = \dfrac{e^{z_k}}{\sum \limits_{i=0}^{K} e^{z_i}}~~~\forall~k=0, \ldots, K
\end{equation}
with $z \in \mathbb{R}^{K+1}$ as the output of the last fully connected layer.
Then the classification loss is defined as the log loss of the true class $y$:
\begin{equation}
    \L_C = -\log(p_y)
\end{equation}

\paragraph[2. Bounding box regression loss $\L_B$]{\it 2. Bounding box regression loss $\L_B$}
The output of the bounding box regression branch is given by a four tuple of pixel values $\hat{b}^k = \left(\hat{b}^k_c, \hat{b}^k_d, \hat{b}^k_w, \hat{b}^k_h\right) \forall~k=1, \ldots, K$. Here,  $(\cdot_c$, $\cdot_d)$ describe the pixel coordinates of the center of the bounding box, while the width and height of the bounding box are given by the pixel values with the indices $w$ and $h$. For detailed information on the derivation of the certain pixel values we refer to \cite{fast_rcnn}. With the ground truth bounding box $b^y = \left(b^y_c, b^y_d, b^y_w, b^y_h\right)$, assigned to each RoI if $y\neq 0$, the loss function is defined by
\begin{equation}
\L_B = \mathbbm{1}_{\{y>0\}}\sum \limits_{j \in \{c, d, w, h \}} \mathcal{H}(\hat{b}^y_j -b^y_j)
\end{equation}
with $\mathcal{H}(\phi)$ the Huber loss function \cite{huber}
\begin{equation}
\mathcal{H}(\phi) = \begin{cases} 0.5 \phi^2, \qquad \, \,  {\rm if}~|\phi|<1\\
|\phi| - 0.5,  \quad {\rm else}
\end{cases}
\end{equation}
and the indicator function
\begin{equation}
\mathbbm{1}_{\{y>0\}}=\begin{cases}
1,\quad {\rm if}~y>0\\
0,\quad {\rm if}~y=0 \, .
\end{cases}
\label{eq:kick_bg}
\end{equation}

\paragraph[3. Loss of the mask branch $\L_M$]{\it 3. Loss of the mask branch $\L_M$}
The output of the mask branch has the dimension $Km^2$ since it encodes a binary mask of a spatial dimension of $m\times m$ for all $K$ object classes except the background class. The values of each pixel of the predicted mask $\hat{\tau}_{rs}^k,~k=1, \ldots, K$ are derived by applying a sigmoid activation function to the outputs of the last feature map. With the pixel values $\tau_{rs}^y$ of the ground truth mask, the loss function is defined as the average binary cross-entropy:
\begin{equation}
\L_M = \mathbbm{1}_{\{y>0\}} \dfrac{1}{m^2} \sum \limits_{r=1}^m \sum \limits_{s=1}^m \tau_{rs}^y \log(\hat{\tau}_{rs}^y ) + (1-\tau_{rs}^y )\log(1-\hat{\tau}_{rs}^y )
\end{equation}
with $\mathbbm{1}_{\{y>0\}}$ defined as in \eqref{eq:kick_bg}.

Those three tasks of classification, bounding box regression and mask generation are solved in the head architecture of the Mask R-CNN which operates on each RoI, whereas the important features are extracted in the convolutional backbone architecture. In this work, the backbone architecture is given by a combination of the ResNet with 101 layers \cite{resnet} and the Feature Pyramid Network (FPN) \cite{fpn}. The head architecture is given only by the FPN.

Summarizingly, the Mask R-CNN generates segmentation masks of instances of interest are segmented from the background and considered individually. 

\section{Datasets for Interior Sensing} \label{sec:data}
Scenarios of interior sensing are  highly complex. The instances separated from the background belong to the classes ``Person'', ``Child seat'' and ``Object'' in this work. Thus, the foreground instances can be both, dynamic and static. Furthermore, only those detected instances which are positioned on the front and back seats of the car are of interest.
Also modelling the background is non-trivial since it contains dynamic elements due to the motion of objects visible through the car windows.
Moreover, the camera position changes slightly from car to car in this work.
Additionally, the impact of environmental effects has to be taken into account, like different weather conditions, shadows, traffic lights and vibrations.

To solve the background-foreground segmentation task in this high complex setting adequately, it is of importance, especially for the training of the Mask R-CNN, that an appropriate training set is available by which a wide range of the challenges is covered. 
To this end, the ISSO dataset has been created by APTIV and the authors of this work. It consists of 1300 real-world images extracted from videos of different interiors of stationary or driving cars. The images contain a high variety regarding the foreground instances, the background and the environmental conditions. Further details are provided in the upcoming section.

Since the annotation of images is time-consuming and costly, only a small amount of 1100 real-world images is available for training. To overcome this problem, we apply transfer learning.  Here, the training of the Mask R-CNN is initialized by a model  pretrained on the COCO dataset. This pretrained model is able to detect persons and certain everyday objects outside the scope of car interiors. Therefore, we also consider the impact of the COCO dataset maintained during the training. Moreover, images of the synthetic dataset SVIRO are used for the training of the Mask R-CNN to overcome the problem of the small amount of real-world annotated data. The SVIRO dataset consists of rendered images describing scenarios in the passenger compartment of different cars.

Hence, in total, we consider images and videos of three different datasets  -- the COCO dataset, the SVIRO dataset and our ISSO dataset. In the next section, we describe these datasets in more detail.

\subsection{The ISSO dataset}
\label{ssec:Aptiv_data}
The Interior Sensing and Seat Occupancy (ISSO) dataset has been created by APTIV and the authors of this work. It consists of images which are extracted from videos that have been recorded in driving or stationary cars by  APTIV in Wuppertal, Germany. The purpose of this dataset is to enable the feasibility study provided by the present article and it is not meant to be representative for a local or global population. While selecting the images for the dataset, it was taken into account, that a high variation within the instances, the backgrounds and the light conditions is given over all images. The camera is mounted on the upper or lower area of the windshield in each car. Hence, the position changes slightly per car. Since the camera is not integrated, its position might even change slightly in one and the same car.
In total, 1300 images were labeled from which we define test, training and validation sets.

\begin{figure}[!h]
    \centering
    \includegraphics[trim =3mm 5mm 3mm 15mm, clip, scale=0.45]{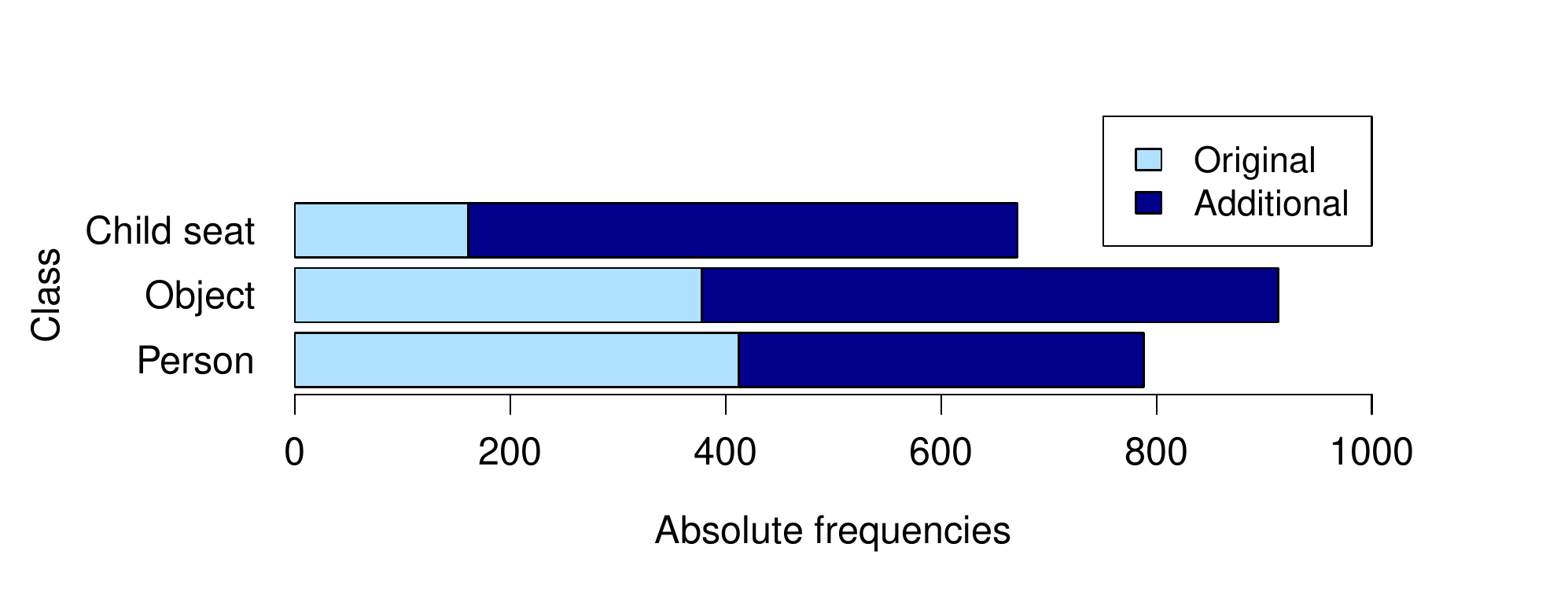}
    \caption{Distribution of the classes over the 1100 images of the training set. By ``Original'' the set of the first 500 images is described. ``Additional'' describes the set of the 600 images by which the training set was extended.}
    \label{fig:test_set}
\end{figure}

\paragraph[1. Training set]{\it 1. Training set}
The training set is used to train the Mask R-CNN. In total, it consists of 1100 labeled images recorded in five different cars. 500 images were selected and labeled at the beginning of this work. Since the class ``Child seat'' suffered from a lack of variation, it was paid attention to recording as many different child seats as possible in later recording session. Of these new videos, 600 additional images were chosen and labeled, such that the number of instances for the class ``Child seat'' increased in particular. Nevertheless, as one can see from \cref{fig:test_set}, the instances of this class are least represented in the training set. For the class ``Child seat'' it is to remark that instances are not clearly visible in two situations. Here, a situation refers to a specific camera position in a specific car. Firstly, due to the camera position, only a small part of a child seat is visible if it is mounted on the back seat of the car interior. Second, a child seat is hardly visible if it is occupied by a child. Hence, especially for the training it is of interest in how many cases the child seats are mounted on the front passenger seat and thus clearly visible.
In particular, the child seats are mounted on the passenger front seat of the car in about $60\%$ of the images that contain a child seat. Of these front mounted child seats over $80\%$ are not occupied. Detailed statistics for the training set are given in the \cref{sec:appendix_aptiv}.

\paragraph[2. Validation set]{\it 2. Validation set}
The validation set consists of 100 images distributed over three different cars. It contains five persons, among them one child, one female and four male. The class ``Object'' is represented by four instances of five main categories (laptop, PC-keyboard, bagpack and beverage crate). Moreover, two child seats are available in the validation dataset. In 29 images, a child seat is contained whereby only seven of these images show a child seat mounted on the passenger front seat. None of the front mounted child seats is occupied by a child. All cars, child seats, objects and persons are different from those shown by the training set.

\paragraph[3. Test set]{\it 3. Test set}
The test set is used to evaluate and to compare the performance of the implemented foreground-background detection methods. It consists of 100 labeled images extracted from 70 videos that are recorded in five different cars. 50 images are extracted from videos inside a driving car and the other 50 images are extracted from videos inside stationary cars.

The test set contains 13 persons, among them three children and one baby, three female and ten male. In the class ``Object'', everyday items are collected, like a bag pack or a wallet. As described in \cref{tab:test_obj}, 42 instances of 16 main categories are included in the test set. Furthermore, four different child seats are available in the test set. The child seat is mounted on the passenger front seat of the car in about $30\%$ of the images that contain a child seat. Additionally, about $53\%$ of these front mounted child seats are occupied. All cars, child seats, objects and persons are different from those shown by the training and validation sets.

\begin{figure}[!h]
\centering
\begin{minipage}{0.23\textwidth}
\includegraphics[width=\textwidth]{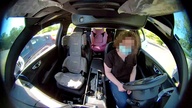}
\end{minipage}%
\hspace{1mm}%
\begin{minipage}{0.23\textwidth}
\includegraphics[width=\textwidth]{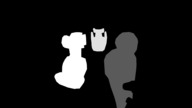}
\end{minipage}
\caption{An image of the training set (left) and its corresponding gray scale ground truth instance segmentation mask (right).}
\label{fig:gtTrain}
\end{figure}

\paragraph[Creation of the ground truth]{\it Creation of the ground truth }
The annotations of the images are created by the tool ``Labelme'' from MIT \cite{labelme} extended by the function of an eraser. With ``Labelme'', it is possible to annotate the instances of an image by closed polygon courses. By this information, the ground truth segmentation masks with different gray scale values for each instance can be created (see \cref{fig:gtTrain}).

\begin{figure}[!h]
\centering
\begin{minipage}{0.23\textwidth}
\includegraphics[width=\textwidth]{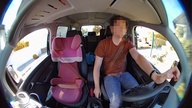}
\end{minipage}%
\hspace{1mm}
\begin{minipage}{0.23\textwidth}
\includegraphics[width=\textwidth]{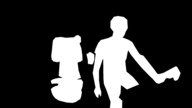}
\end{minipage}
\caption{An image of the test set (left) and its corresponding binary ground truth segmentation mask (right).}
\label{fig:gtTest}
\end{figure}

In this feasibility study we only focus on foreground-background segmentation. Hence, for the images of the test set, we generated binary ground truth segmentation masks. As described by \cref{fig:gtTest}, the foreground instances are represented by white pixels, while the background is given by black pixels. 
\vspace{-5mm}

\subsection{The SVIRO dataset}
As the name suggests, the Synthetic Vehicle Interior Rear Seat Occupancy (SVIRO) dataset \cite{sviro} consists of images produced artificially by a rendering software which is Blender, version 2.79. These images depict randomly generated scenarios in the passenger compartment of ten different vehicles. For each of these cars 2500 images were generated and split into a training and a test set. Here, each training set contains 2000 labeled images and each test set 500 images.
The labeled instances of the classes ``Person'', ``Child seat'' and ``Everyday object'' differ between the training and the test set.

\begin{figure}[!h]
\centering
\begin{minipage}{0.22\textwidth}
\includegraphics[width=\textwidth]{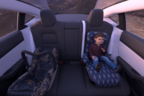}
\end{minipage}
\begin{minipage}{0.22\textwidth}
\includegraphics[width=\textwidth]{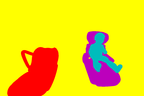}
\end{minipage}
\caption{An RGB image (left) and the corresponding ground truth instance segmentation mask (right) of the SVIRO dataset. Source: \cite{sviro}.}
\label{fig:sviro}
\end{figure}

Herein, the training datasets of the five car models Ford Escape, the Lexus GSF, the Tesla Model 3, the VW Tiguan and the RenaultZoe are used. For the training set of each car, the following statistics apply:
Each training set contains 23 persons, whereby six persons are children and three persons are babies. Moreover, three different child seats and four different everyday objects are used in one training dataset. Furthermore, different light conditions are taken into account.
For the task of instance segmentation, the ground truth corresponding to the RGB image is given by an instance segmentation mask as described in \cref{fig:sviro}.

\subsection{The COCO dataset}
\label{ssec:coco}
The ``Common Objects in Context'' (COCO, \cite{coco}) dataset
consists of images that depict everyday objects in typical environments. Mainly, the images are non-iconic. This means for example, that an object is not  shown in front of a calm background but in a complex scene.
For this work the training dataset of the year 2017 is used. This dataset consists of over 118,000 labeled images covering over 60 object categories of 10 main categories including the category ``background'' as described in \cref{tab:appendix_coco}. The annotations are created by labeling each instance in an image by a closed polygon course and their corresponding bounding box.

\section{Evaluation Metrics}
\label{sec:evaluation}
The foreground-background masks generated by the implemented frameworks are evaluated pixel-wise. The metrics by which this evaluation is realized are composed of the following four terms: 
\begin{itemize}
\item True positives (TP): Pixels that belong to the foreground and which are correctly classified.
\item False positives (FP): Pixels that belong to the background, but which are misclassified. 
\item  True negatives (TN): Pixels that belong to the background and which are correctly classified.
\item False negatives (FN): Pixels that belong to the foreground, but which are misclassified. 
\end{itemize}
Thereof, commonly used metrics for the evaluation of background-foreground segmentation tasks \cite{all_metrics} can be defined as given in \cref{tab:evaluation_metrics_def}.
While the precision describes the amount correctly predicted foreground pixels relative to the total number of predicted foreground pixels, recall considers the predicted foreground pixels relative to the total number of actual (true) foreground pixels, corresponding to the ground truth.
The specificity describes the proportion of true background pixels that are correctly classified. The accuracy provides the proportion of correct classifications overall.
The similarity is also known as the Jaccard index or the Intersection over Union (IoU) \cite{all_metrics, jacc, iou}. This value measures to what extent the ground truth mask and the predicted mask resemble one another.
Finally, the $F_1$-score describes the  harmonic mean of precision and recall \cite{fbeta}. 
\begin{table}[!h]
    \centering
    \begin{tabular}{|m{1.8cm}|m{5cm}|}
    \hline
    \rowcolor{bblue} Name &  Formula  \\ \hline
     Precision  &  $Pr = \frac{TP}{TP+FP}$ \Tstrut\Bstrut \\ \hline
    Recall & $Re = \frac{TP}{TP+FN}$ \Tstrut\Bstrut \\ \hline
      Specificity &  $Sp = \frac{TN}{TN+FP}$ \Tstrut\Bstrut \\ \hline
      Accuracy &  $Acc = \frac{TP+TN}{TP+TN+FP+FN}$ \Tstrut\Bstrut \\ \hline
     Similarity &  $Sim = \frac{TP}{TP+FP+FN}$ \Tstrut\Bstrut \\ \hline
     $F_{1}$-score & \ $F_{1} = \frac{Pr\cdot Re}{Pr +Re} = \frac{2TP}{2TP+FP+FN} $ \Tstrut\Bstrut \\ \hline
    \end{tabular}
    \vspace{1mm}
    \caption{Definition of the evaluation metrics.}
    \label{tab:evaluation_metrics_def}
\end{table}

\begin{figure}[!h]
\centering
\includegraphics[trim=0mm 0mm 0mm 10mm, clip, scale=0.55]{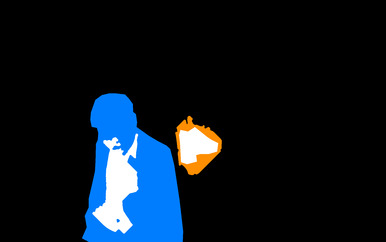}
\caption{A  predicted foreground-background mask. Here, the colored pixels are coded as follows: TP = white pixels, FP = orange pixels, TN = black pixels, FN = blue pixels.}
\label{fig:metric_pred_mask}
\end{figure}

\begin{remark}{{\bf (Evaluation on images of test set)}}\\
The performance of all methods is evaluated on the test set presented in \cref{ssec:Aptiv_data}. In section \cref{sec:results}, the averaged values over the 100 images are reported for all evaluation metrics.
Furthermore, in our comparisons we mainly rely on the metric ``similarity'' which we compute for each test image separately and then compute the average over the whole test set.
The similarity is the most sensitive and intuitive metric to measure the differences between the ground truth and the predicted mask. As can be seen in \cref{fig:metric_pred_mask}, the background of the given scenarios is very dominant in relation to the foreground instances. In this case, the accuracy would be still on a high level, if the foreground instances would not be segmented properly or detected at all. Due to this reason, the accuracy can lead to misinterpretations of the results. Contrary, the similarity is not affected by the distribution of background and foreground pixels over a given image. So, this metric produces more robust results. For example, for the predicted mask shown in \cref{fig:metric_pred_mask}  the similarity is only $26\%$ but the accuracy is about $90\%$. In section \cref{sec:results} the similarity of the prediction and ground truth masks is denoted by the term ``similarity score''.
\end{remark} 
\vspace{-3mm}

\section{Results of Experiments}
\label{sec:results}
For each of the methods introduced in \cref{sec:methods}, implementation details and the results of our experiments are summarized in the following. All codes have been written in Python.

\subsection{Gaussian Mixture Model (GMM)}
\paragraph[Implementation details]{\bf Implementation details}
For the implementation of the GMM, the OpenCV-function {\tt BackgroundSubtractorMOG2} \cite{gmm_opencv, gmm_opencv_docu} implemented in Python was used.  By this function, a background model can be created. Contrary to the other two considered methods, the input of the GMM has to be a video since it is a motion based approach. Thus, the GMM was applied on the entire videos, from which the predicted masks of the test images were extracted.

To initialize the background model, mainly one parameter has to be set, namely $T=\tau N$ (recall \cref{ssec:GMM}, $N$ denotes the number of previous frames to be considered). This parameter indicates, the number of the last frames which are affecting the background model. After parameter tuning, we fixed the value $T=250$ for all  presented experiments below. Furthermore,
the algorithm choses a learning rate $\mathit{lr} \in (0,1) $ automatically where $\mathit{lr}=0$ would mean that the background model is never updated and $\mathit{lr}=1$ would mean that the background model is newly initialized for every frame. For further details, see \cite{gmm_opencv_apply}.

\begin{figure}[!h]
    \centering
    \includegraphics[scale=0.22]{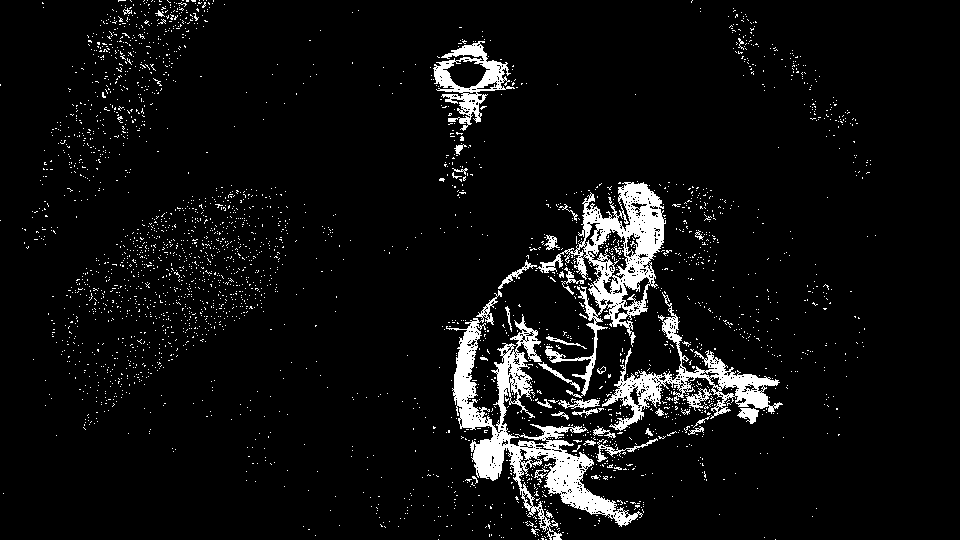}
    \caption{A foreground-background mask predicted by the GMM.}
    \label{fig:gmm_example}
\end{figure}

\paragraph[Results]{\bf Results}
The masks predicted by the GMM achieve similarity of $15.5\%$ (averaged over the whole test set). We investigated if the quality of the masks can be enhanced by the application of a post-processing and pre-processing method. In particular, one can observe, that the predicted foreground-background masks contain a lot of noise as described by \cref{fig:gmm_example}. Furthermore, the boundaries of the detected instances are clearly visible,  however big parts of the areas inside the instance boundaries are not predicted as foreground. Therefore, we investigated the effect of the morphological operators ``Opening'' and ``Closing''.
Both operators are defined by a composition of the morphological operators Dilation and Erosion presented in \cref{sec:morph_snakes}. While noise can be removed from binary segmentation masks by the operator Opening, the areas of the instance boundaries can be filled by white pixel values with the help of the operator Closing.
\begin{remark}{\rm \bf (Notation)}\\
In the subsequent tables, the columns ``MO'', `C' and ``CE'' contain the following information:
\begin{itemize}
    \item MO describes which morphological operator was applied: C = Closing and O = Opening.
    \item C shows if the color channel V or L was considered.
    \item CE represents the applied contrast enhancement method: HE (Histogram equalization) or CHE (CLAHE) (see \cref{ssec:color}).
    \item The baseline is given by the model with no application of a pre- or post-processing marked by the entry ``-'' of the columns MO, C and CE.
\end{itemize}
\end{remark}
\begin{table}[!h]
\centering
\begin{tabular}{|c|cccccc|}
 \hline
MO & $Pr$ & $Re$ & $Sp$ & $Acc$ & $Sim$ & $F_1$ \\
 \hline
-  & 0.333 & 0.325 & 0.811 & 0.744 & 0.155 & 0.256  \\ 
\hline
\rowcolor{lgray1}
 C  & 0.337 & 0.368 & 0.794 & 0.736 & {\bf 0.172} & 0.279 \\ 
 O  & 0.357 & 0.269 & 0.837 & 0.756 & 0.131 & 0.220  \\ 
  \hline
\end{tabular}
\vspace{1mm}
\caption{Evaluation results for the predicted masks of the GMM after applying post-processing methods.}
\label{tab:gmm_post}
\end{table}

\begin{table}[!h]
\centering
\resizebox{8.5cm}{!}{
\begin{tabular}{|c|c|cccccc|}
 \hline
C & CE & $Pr$ & $Re$ & $Sp$ & $Acc$ & $Sim$ & $F_1$ \\ 
 \hline
-  &  -  & 0.333 & 0.325 & 0.811 & 0.744 & 0.155 & 0.256  \\ 
\hline
 V  &  HE  & 0.289 & 0.380 & 0.776 & 0.720 & 0.163 & 0.268 \\ 
  \rowcolor{lgray1}
 V &  CHE  & 0.309 & 0.434 & 0.770 & 0.725 &{\bf  0.192} & 0.307 \\ 
 \hline
 L&  HE  & 0.298 & 0.351 & 0.808 & 0.744 & 0.161 & 0.267 \\ 
  \rowcolor{lgray1}
 L &  CHE  & 0.315 & 0.418 & 0.785 & 0.736 &{\bf  0.192} & 0.308  \\ 
  \hline
\end{tabular}
}
\vspace{1mm}
\caption{Evaluation results for the predicted masks of the GMM generated of preprocessed frames.}
\label{tab:gmm_pre}
\end{table}

\begin{table}[!h]
\centering
\resizebox{8.5cm}{!}{
\begin{tabular}{|c|c|cccccc|}
 \hline
MO & C & $Pr$ & $Re$ & $Sp$ & $Acc$ & $Sim$ & $F_1$ \\
 \hline
-  &  -  & 0.333 & 0.325 & 0.811 & 0.744 & 0.155 & 0.256  \\ 
\hline
\rowcolor{lgray1}
 C  & V & 0.311 & 0.503 & 0.740 & 0.710 & {\bf 0.212} & 0.332 \\ 
 O  & V & 0.345 & 0.358 & 0.815 & 0.750 & 0.175 & 0.282 \\ 
  \hline
  \rowcolor{lgray1}
 C & L  & 0.319 & 0.492 & 0.754 & 0.722 & {\bf  0.214} & 0.336 \\ 
 O & L & 0.350 & 0.335 & 0.831 & 0.762 & 0.171 & 0.278 \\ 
  \hline
\end{tabular}
}
\vspace{1mm}
\caption{Evaluation results for the predicted masks of the GMM after applying post-processing methods on the masks. The input images are pre-processed by applying CLAHE on the respective color channel.}
\label{tab:gmm_all}
\end{table}
As  shown by \cref{tab:gmm_post}, the application of Closing on the predicted masks leads to a slight improvement of about $2$ percent points (pp.) for the similarity, whereas the quality of the predicted masks decreases when removing noise via Opening. Moreover, the performance of the GMM suffers from changes in the light conditions. To address this problem, a pre-processing step $P_{CC}$ was introduced. Here, an image of the RGB color space is converted into the color spaces HSV or Lab. Then, the respective color channels V or L are extracted since it is possible to control the brightness of an image by them as described in \cref{ssec:color}. Additionally, Histogram equalization or CLAHE can be applied to the extracted channels to approximate a uniform distribution regarding the brightness of an image and to enhance the contrast at the same time. Thus, each frame of a video sequence was pre-processed by $P_{CC}$ before the GMM algorithm was applied on it. The corresponding results are given in \cref{tab:gmm_all}.
We observe that the similarity score increases slightly by about $4$ pp.\  by applying CLAHE on the color channels V or L. If this pre-processing method was applied together with the morphological operator Closing as post-processing, the similarity even reached a value of about $21\%$ as presented in \cref{tab:gmm_all}.

\paragraph[Limitations]{\bf Limitations}
The prediction of foreground-background masks by the GMM approach poses problems for the given scenario. The main reason is that the GMM is a motion based approach. The detection of static foreground instances that primarily belong to the classes ``Object'' or ``Child seat'' depends highly on external effects as the movement by a person or by the vibrations during the drive. But also foreground instances, which are dynamic in general, might not be detected if they do not move over a longer time period, such that they can be incorporated to the background.
Additionally, a high number of false positives is generated when the car is driving. In our experiments, we observed that this can be attributed to motion visible in the car windows but also changing illumination.
Moreover, the GMM takes a while to learn the background model. Hence, predicted masks for frames at the beginning are of bad quality.

\subsection{Morphological Snakes}
\paragraph[Implementation details]{\bf Implementation details}
To implement the morphological snakes, the Python package  {\tt morphsnakes} \cite{pmneila_morphsnakes} was used. The input of the algorithm is a gray scale image. For the MACWE as well as for the MGAC the initial contour is given by a circle for which the center $(x,y)$ and the radius $r$ have to be determined by the user. Here, a contour was initialized on each car seat which is occupied by at least one foreground instance. Furthermore, the number of iterations $i$ for the curve evolution and the number of smoothing steps $s \in \{1, 2, 3, 4\}$ have to be defined for both methods, here $s=2$. Besides that, the following parameters have  been set for each method individually.

\paragraph[1. MACWE]{\it 1. MACWE}
For the MACWE, the weight parameters for the region outside $\lambda_1$ and inside $\lambda_2$ the evolving curve have to be defined. In the case $\lambda_1 > \lambda_2$, it is  assumed that the region outside the evolving curve contains more variation in their pixel values, compared to the region inside the curve and vice versa.

\begin{figure}[!h]
    \centering
    \includegraphics[scale=0.45]{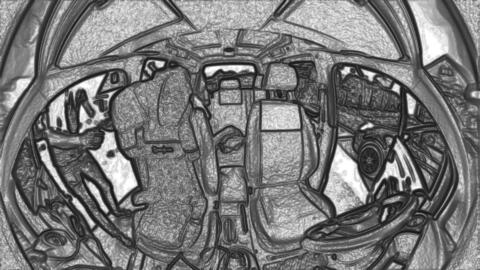}
    \caption{A gray scale real-world image after the application of the pre-processing $P_{\rm MGAC}$.}
    \label{fig:mgac_pre}
\end{figure}

\paragraph[2. MGAC]{\it 2. MGAC}
To perform the MGAC algorithm properly on an image, the contours of the foreground instances need to be clearly visible. Due to this reason, a pre-processing $P_{\rm MGAC}$ was performed on the images to highlight these contours. Here, $P_{\rm MGAC}$ is given by an ``inverse gaussian gradient magnitude''-filter defined in \eqref{eq:mgac_filter}. 
Applying this filter  to an image leads to  all pixel values  being inside the interval $[0,1]$. By this, the pixel values are close to zero, particularly in the areas which are close to the contours of the foreground instances as shown in \cref{fig:mgac_pre}.
To conduct this pre-processing, two parameters have to be defined. The standard deviation $\sigma$ of the Gaussian filter and the non-linear scaling parameter $\alpha$ acting as a steepness parameter. The larger $\alpha$ is, the steeper the transition between the areas of the instance contours and the flat regions inside and outside their contours is. Here, $\sigma=3$ and $\alpha=1000$.

Additionally, two parameters have to be set for the actual MGAC algorithm. The balloon force term $v \in \mathbb{R}$ determines if a Dilation ($v>0$) or Erosion ($v<0$) should be performed. By $v=0$, no balloon force is applied. As a result of parameter tuning, we set the value to $v=1.2$. Secondly, the stopping threshold $\tau$ has to be defined. Regions of the image with smaller values than $\tau$ are considered as the contours of the foreground instances. Thus, the evolution of the curve stops in those regions. In the experiments, we consider different values of $\tau$.

\paragraph[Results]{\bf Results}

\paragraph[1. MACWE]{\it 1. MACWE}
We study the influence of the number of iterations $i$ and of the weights parameters $\lambda_1$, $\lambda_2$. In particular, three different values are considered for $i$, $\lambda_1$ and $\lambda_2$ respectively, namely $i \in \{100, 200, 300\}$ and $\lambda_1$, $\lambda_2$ $\in \{1,2,3\}$. The baseline is represented by the assumption, that the pixel values of the regions inside and outside the evolving curve contain the same amount of variation, so $\lambda_1 = \lambda_2 = 1$.

As  shown by \cref{tab:MACWE}, the similarity score increases with the increase of $\lambda_2$ for the case $\lambda_2 \geq \lambda_1=1$. During the experiments, we observed that the number of false positives increases by the increase of iterations $i$. Therefore, the values for the precision, accuracy, similarity and $F_1$-score decrease with the increase of $i$.  We made the same observation for the case $\lambda_1 \geq \lambda_2=1$. Here, the results of the baseline could not be exceeded for  any $i$. Therefore, the  hypothesis that the variation in the pixel values of the region outside the evolving the curve is higher compared to the pixel values inside the curve can be rejected for the given test set.

\begin{table}[!h]
\centering
\resizebox{8.5cm}{!}{
\begin{tabular}{|c|c|cccccc|}
\hline 
$i$ & $\lambda_2$ & $Pr$ & $Re$ & $Sp$ & $Acc$ & $Sim$ & $F_1$ \\
\hline
100 & 1 & 0.810 & 0.343 & 0.989 & 0.892 &  0.313 & 0.465 \\
100 & 2 & 0.762 & 0.400 & 0.984 & 0.896 & 0.345 & 0.501 \\
\rowcolor{lgray1}
100 & 3 & 0.740 & 0.437 & 0.980 & 0.897 & {\bf 0.366} & 0.525  \\
\hline
200 & 1 & 0.671 & 0.384 & 0.970 & 0.882 & 0.315 & 0.463 \\
200 & 2 & 0.616 & 0.455 & 0.956 & 0.880 & 0.338 & 0.491 \\
\rowcolor{lgray1}
200 & 3 & 0.588 & 0.514 & 0.943 & 0.877 & 0.358 & 0.513 \\
\hline
300 & 1 & 0.593 & 0.406 & 0.951 & 0.870 & 0.309 & 0.454 \\
300 & 2 & 0.533 & 0.484 & 0.927 & 0.860 & 0.323 & 0.471 \\
\rowcolor{lgray1}
300 & 3 & 0.506 & 0.558 & 0.905 & 0.852 & 0.340 & 0.491 \\
\hline
\end{tabular}
}
\vspace{1mm}
\caption{Evaluation results for MACWE with $\lambda_1 = 1$ fixed.}
\label{tab:MACWE}
\end{table}

\paragraph[2. MGAC]{\it 2. MGAC}
Analogously to the MACWE, the influence of the number of iterations and the stopping threshold on the performance of MGAC were investigated. In particular, the values $\tau \in \{0.1, 0.2, 0.3, 0.4, 0.5\}$ are considered for each number of iterations $i \in \{100, 200,  300\}$ in the experiments.
\Cref{tab:MGAC} does not show a clear trend regarding the influence of the number of iterations for MGAC. For each $i \in \{100, 200,  300\}$ we observe that the values of $Sim$, $Acc$ and $F_1$ increase with increasing values of $\tau$ until the respective best values are obtained. For $\tau > 0.4$ we observe in our experiments that the similarity, accuracy and $F_1$-score decrease with increasing $\tau$. 

\begin{table}[!h]
\centering
\resizebox{8.5cm}{!}{
\begin{tabular}{|c|c|cccccc|}
 \hline
$i$ & $\tau$ & $Pr$ & $Re$ & $Sp$ & $Acc$ & $Sim$ & $F_1$ \\ 
 \hline
 100 &  0.3 & 0.788 & 0.572 & 0.981 & 0.914 &  0.469 & 0.633  \\ 
 200 &  0.4 & 0.778 & 0.565 & 0.975 & 0.911 & 0.466 & 0.627 \\ 
 300 &  0.4 & 0.712 & 0.603 & 0.958 & 0.902 & 0.456 & 0.616  \\ 
  \hline
\end{tabular}
}
\vspace{1mm}
\caption{Best evaluation results for each $i$ for MGAC.}
\label{tab:MGAC}
\end{table}

\paragraph[Limitations]{\bf Limitations}
Although the similarity score of the MGAC is about $10$ pp.\ higher than the score of MACWE, both approaches have the same limitations.
The generated masks lose in quality if the pixel values of the foreground instances and the background have similar values in RGB color space. Furthermore, even slight changes in the light conditions and shadows lead to poorly generated foreground-background masks. One can  observe that the curve evolution quickly gets stuck in areas with strong sunlight or at  boundaries of shadows. The performance of the approaches depends highly on exterior factors since the essential parameters have to be determined by the user at the beginning. Hence, the configuration of the algorithms depends on the experience of the user.

\paragraph[Improvement of performance]{\bf Improvement of performance}
Analogously  to  the  GMM,  the  performance  of  the  morphological snakes suffers  from  changes  in  the  light conditions.  For  this  reason  we  also  investigated  if  the  performance could be positively affected by the application of the pre-processing $P_{CC}$.
Thus, instead of a general gray scale image, the contrast enhanced image of the color channels V or L is presented to the respective morphological algorithm.

We repeated the experiments with the best results for both methods, MACWE and MGAC, now incorporating the $P_{CC}$ pre-processing.
The MACWE approach was repeated with the parameters $\lambda_1=1$ and $\lambda_2=3$ for all  $i \in \{100, 200, 300\}$. The MGAC approach was repeated for the combination of the parameters, which are recorded in \cref{tab:MGAC} whereby the pre-processing $P_{CC}$ was performed before $P_{\rm MGAC}$.
Especially for the MACWE, the similarity of the generated and the ground truth masks increased by up to $6$ pp.\  by applying HE to V. The similarity of the masks generated by the MGAC approach and the ground truth masks increased only slightly by $2.05$ pp.

Overall, the highest accuracy ($91.86\%$), $F_1$-score ($64.76\%$) and similarity ($48.63\%$) was achieved by applying MGAC with the parameters $i=100$ and $\tau=3$ on the pre-processed image of the color channel V whose contrast was enhanced by using histogram equalization. 

\subsection{Mask R-CNN}
\paragraph[Implementation details]{\bf Implementation details}
For the experiments with the Mask R-CNN, we used the implementation from \cite{matterport-maskrcnn-2017}.
Furthermore, we applied transfer learning, i.e., we trained the Mask R-CNN by using the weights of a pre-trained model as initial weights. This pre-trained model was trained on the entire COCO dataset presented in \cref{ssec:coco}.
The training of the Mask R-CNN was performed on one Titan XP GPU with 12 gigabytes working memory over 100 epochs with a batch size of 1. Per epoch, 1000 gradient descent steps were performed. Here, the momentum of the adam optimizer \cite{kingma2017adam} was  fixed to 0.9.
During the inference, the detection of an instance was accepted if the predicted probability was $\geq 0.9$. The results of the experiments below are given for a confidence level of $90\%$.
Within our experiments, we study the influence of three factors on the performance of the model:

\paragraph[1. The influence of the learning rate $\mathit{lr}$ and the weight decay $\lambda$] {\it 1. The influence of the learning rate $\mathit{lr}$ and the weight decay $\lambda$} Here, the values $\mathit{lr} \in \{0.0005, 0.001, 0.002, 0.01\}$ and $\lambda \in \{0.0001, 0.001, 0.01\}$ were considered.

\paragraph[2. The influence of data augmentation]{\it 2. The influence of data augmentation} The goal of data augmentation is to increase variability of a dataset. This of particular interest when the dataset is small. Since the ISSO training dataset consists only of 1100 annotated real-world images, we consider data offline augmentation (before training) and online augmentation (during training).
We utilize both online and offline augmentation since different kinds of augmentation are readily available in python. 

To this end, each image and its corresponding ground truth mask were horizontally flipped, randomly cropped and as a combination of both stored to the offline augmented data set.
Additionally, we utilize further augmentation methods in an online augmentation pipeline. To this end, we use the
Python package {\tt albumentations} \cite{onAug}. In detail, the pipeline performs four steps, in each step randomly choosing one of the following augmentation methods:
\begin{enumerate}
\item Gaussian blur, glass blur;
\item Gaussian noise, ISO noise;
\item Random brightness contrast, CLAHE (see \cref{ssec:color}), random sunflair;
\item Grid distortion, elastic transformation, optical distortion.
\end{enumerate}
In particular, the methods of (4) affect also the appearance of the ground truth masks.

\paragraph[3. The influence of the data which is used during training]{\it 3. The influence of the data which is used during training} 
\label{para:mrcnn_influence_data}
The Mask R-CNN was trained on the three different datasets presented in \cref{sec:data}. 
Below, the datasets used in the training are denoted as follows:
\begin{itemize}
\item $\mathrm{A}^{j}$: The ISSO training dataset which contains $j$ real-world images, $j \in \{500, 1100\}$.
\item $\mathrm{A}^{j}_{\rm aug}$: The ISSO training dataset which contains additionally the images of the offline augmentation. Thus, in total, this dataset consists of $4j$ images, $j \in \{500, 1100\}$.
\item $\mathrm{S}^{j}$: A subset of the SVIRO training set. This subset consists of $j$ images of five different car interiors of the SVIRO dataset, $j \in \{2000, 4400\}$. 
The number of images of each car is uniformly distributed in the dataset $\mathrm{S}^j$.
\item $\mathrm{C}^{j}$, $j \in \{4000, T\}$: A subset of the COCO dataset. The subset $\mathrm{C}^T$ consists of all images, which do not belong to the main category ``vehicle'', ``outdoor'' or ``animal'' (see \cref{tab:appendix_coco}). That is, $\mathrm{C}^T$ contains images illustrating persons and everyday objects. The instances of all main categories are summarized to the class ``Object'', except the instances of the main category ``Person''.  $\mathrm{C}^{4000}$ consists of 4000 randomly sampled images of $\mathrm{C}^T$. 
\end{itemize}
Lastly, the ISSO validation set was used for monitoring the training progress and tuning parameters.

\paragraph[Results]{\bf Results}

\begin{table*}
\centering
\resizebox{10cm}{!}{
\begin{tabular}{|c|cccccc|}
\hline
Data & $Pr$ & $Re$ & $Sp$ & $Acc$ & $Sim$ & $F_1$ \\ \hline
$\mathrm{A}^{500}_{\rm aug}$ + $\mathrm{S}^{2000}$ & 0.904 & 0.790 & 0.989 & 0.965 &  0.735 & 0.840 \\ \hline
$\mathrm{A}^{500}$ & 0.904 & 0.752 & 0.990 & 0.961 & 0.705 & 0.822  \\ 
$\mathrm{A}^{500}_{\rm aug}$ & 0.939 & 0.736 & 0.991 & 0.960 & 0.700 & 0.805  \\ 
$\mathrm{S}^{2000}$ & 0.630 & 0.659 & 0.953 & 0.925 & 0.520 & 0.737 \\ \hline
$\mathrm{A}^{500}_{\rm aug}$ + $\mathrm{C}^{T}$+$\mathrm{S}^{2000}$ & 0.739 & 0.717 & 0.967 & 0.940 & 0.594 & 0.742  \\
$\mathrm{A}^{500}_{\rm aug}$ + $\mathrm{C}^{4000}$+$\mathrm{S}^{2000}$  & 0.863 & 0.782 & 0.985 & 0.962 &  0.708 & 0.813 \\
$\mathrm{A}^{500}_{\rm aug}$ + $\mathrm{C}^{4000}$  & 0.837 & 0.778 & 0.979 & 0.958 & 0.680 & 0.788  \\ \hline
\rowcolor{lgray1}
$\mathrm{A}^{1100}$ & 0.919 & 0.797 & 0.991 & 0.969 & {\bf 0.755} & 0.866  \\
$\mathrm{A}^{1100}_{\rm aug}$  & 0.888 & 0.759 & 0.988 & 0.962 & 0.705 & 0.813  \\ 
$\mathrm{A}^{1100}_{\rm aug}$ + $\mathrm{S}^{4400}$  & 0.804 & 0.801 & 0.967 & 0.947 & 0.669 & 0.778 \\
\hline
\end{tabular}
}
\vspace{1mm}
\caption{Evaluation results for models trained on different datasets by applying online augmentation with the parameters $lr=0.001$ and $\lambda=0.0001$.}
\label{tab:mrcnn_data}
\end{table*}  

Firstly, we discuss the results of experiments that were obtained from the Mask R-CNN trained on the datasets  $\mathrm{A}^{500}_{\rm aug}$ as well as $\mathrm{S}^{2000}$ (4000 images in total) and consider the effect of online augmentation. Offline augmentation is performed by default and consists of different augmentations than the online augmentation. The learning rate and weight decay were adopted from \cite{mrcnn}. As documented by \cref{tab:mrcnn_onoff} the similarity is about $4$ pp.\ higher if additional online augmentation is performed. Likewise, the precision also increases. Due to this reason, we applied online augmentation in all subsequent experiments.

Next, the influence of the learning rate and the weight decay are considered. To test the influence of the parameter $\mathit{lr}$ the weight decay was set to $\lambda=0.0001$.
\Cref{tab:mrcnn_lr} suggest that $\mathit{lr=0.001}$ is a descent choice. For a higher learning rate the final similarity score after training decreases. 
In all subsequent experiments, we fixed $lr=0.001$.
In order to test the influence of the weight decay we increased $\lambda$ and considered $lr=0.001$ and $lr=0.01$. This results in similarity scores of $69.61\%$ and $70.58\%$, respectively, therefore neither showing a clear tendency nor a performance increase. Hence, we fix $\lambda=0.0001$.

\begin{table}[!h]
\centering
\resizebox{8.5cm}{!}{
\begin{tabular}{|c|cccccc|}
\hline 
Online aug.  & $Pr$ & $Re$ & $Sp$ & $Acc$ & $Sim$ & $F_1$ \\
\hline
No & 0.830 & 0.762 & 0.982 & 0.960 & 0.677 & 0.801 \\ 
\rowcolor{lgray1}
Yes & 0.920 & 0.767 & 0.989 & 0.961 & {\bf 0.716} & 0.817 \\ 
\hline
\end{tabular}
}
\vspace{1mm}
\caption{Evaluation results for models trained on $\mathrm{A}^{500}_{\rm aug}$ and $\mathrm{S}^{2000}$ with $lr=0.002$ and $\lambda=0.0001$.}
\label{tab:mrcnn_onoff}
\end{table}

\begin{table}[!h]
    \centering
    \resizebox{8.5cm}{!}{
    \begin{tabular}{|c|cccccc|}
    \hline 
    $lr$ & $Pr$ & $Re$ & $Sp$ & $Acc$ & $Sim$ & $F_1$ \\
    \hline
    0.002 &  0.920 & 0.767 & 0.989 & 0.961 &  0.716 & 0.817  \\
    \rowcolor{lgray1}  
    0.001 & 0.904 & 0.790 & 0.989 & 0.965 & {\bf 0.735} & 0.840 \\ 
    0.0005 & 0.894 & 0.752 & 0.987 & 0.961 & 0.694 & 0.800 \\ 
    0.01 & 0.000 & 0.000 & 1.000 & 0.855 & 0.000 & NA  \\ 
    \hline
    \end{tabular}
    }
    \vspace{1mm}
    \caption{Evaluation results for models trained on $\mathrm{A}^{500}_{\rm aug}$ and 
    $\mathrm{S}^{2000}$ with $\lambda =0.0001$.}
    \label{tab:mrcnn_lr}
\end{table}
In addition to the experiments on data augmentation, we consider different compositions of training sets and study their influence on the similarity score after training. The results are summarized in \cref{tab:mrcnn_data}.
For this discussion, the previously best model trained on $\mathrm{A}^{500}_{\rm aug}$ and $\mathrm{S}^{2000}$ with a similarity of $73.5\%$ sets the baseline. As described in \cref{tab:mrcnn_data}, the Mask R-CNN trained only on the real-world images of the ISSO dataset $\mathrm{A}^{500}$ or $\mathrm{A}^{500}_{\rm aug}$ reaches a similiarity of about $70\%$. A model solely trained on the synthetic data $\mathrm{S}^{2000}$ performs significantly worse, achieving a similarity score of $52.0\%$. This signals the presence of a strong domain shift when going from synthetic to real data which is typical for machine learning in computer vision \cite{domainShift}.
Since the initial weights were pretrained on the whole COCO dataset, we study if the model's performance can be  improved by adding the subset $C^T$ or $C^{4000}$ to the training. This approach aims at making the model memorize at least some of the features from the COCO dataset. Contrary to the ISSO and the SVIRO dataset, the images of the COCO dataset show the foreground instances not in the setting of car interiors but in an arbitrary environment. In \cref{tab:mrcnn_data} it can be observed that the performance of the model decreases significantly, compared to the baseline, if the number of images of the COCO dataset is much bigger than the number of images describing the foreground instances in the setting of car interiors during the training. Only the model trained on a balanced dataset with 4000 images of the COCO dataset ($C^{4000}$) and 4000 images of the ISSO and SVRIO dataset ($\mathrm{A}^{500}_{\rm aug}$ $+$ $\mathrm{S}^{2000}$) reaches a similarity of about $71\%$. Nonetheless, the model does not outperform the baseline.

When studying the predicted segmentations per image, the problem observed for the baseline model is that child seats are not detected  reliably. This instability might occur due to the small variation of four child seats in the datasets $\mathrm{A}^{500}$ and $\mathrm{A}^{500}_{\rm aug}$. To test this hypothesis the Mask R-CNN was trained another time (after an extensions of the data collection process) by the extended datasets $\mathrm{A}^{1100}$ and $\mathrm{A}^{1100}_{\rm aug}$ which contain 16 different child seats in total. To balance the ratio between synthetic and real-world data, the amount of images of the SVIRO dataset was also enlarged for the training.
\Cref{tab:mrcnn_data} shows that the similarity increases by training the Mask R-CNN just on the 1100 real-world images of the extended dataset. By looking through the background-foreground masks predicted by this model we observed that all unoccupied child seats are clearly segmented. However, occupied ones still remain challenging.

In conclusion, the best model with a similarity score of $75.5\%$ was obtained by training the model on the dataset $\mathrm{A}^{1100}$ with the use of online augmentation and the determination of the parameters $lr=0.001$ and $\lambda=0.0001$. 

\paragraph[Limitations]{\bf Limitations}
The detection of occupied child seats and instances on the back seats of the car represent still difficult cases. As described in \cref{ssec:Aptiv_data}, only a small number of occupied child seats is given in the ISSO training dataset. Hence, this problem might be solved analogously to the problem with the unoccupied child seats by extending the dataset by more images of occupied child seats. The problem of the detection on the back seats could be solved by additional cameras, such that the instances on the back seats become sufficiently visible for the detection task. 

\subsection{Comparison}
\begin{figure*}
    \centering
    \begin{minipage}{0.24\textwidth}
    \includegraphics[width=\textwidth]{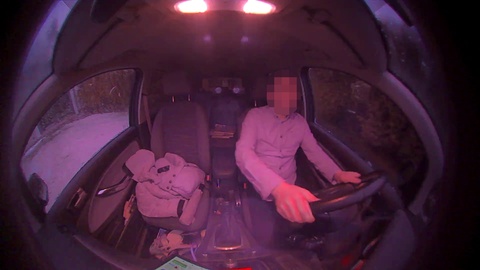}
    \end{minipage}
    \begin{minipage}{0.24\textwidth}
    \includegraphics[width=\textwidth]{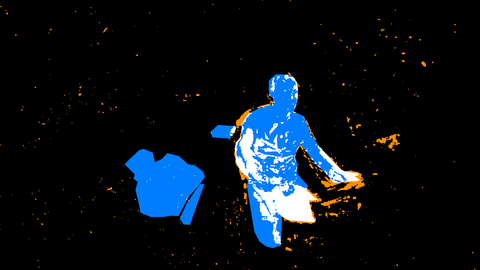}
    \end{minipage}
    \begin{minipage}{0.24\textwidth}
    \includegraphics[width=\textwidth]{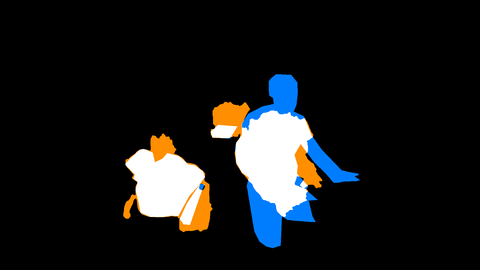}
    \end{minipage}
    \begin{minipage}{0.24\textwidth}
    \includegraphics[width=\textwidth]{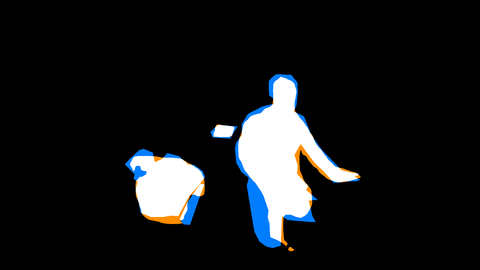}
    \end{minipage}\\[6pt]
    
    \begin{minipage}{0.24\textwidth}
    \includegraphics[width=\textwidth]{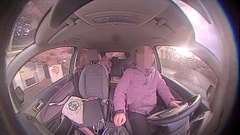}
    \end{minipage}
    \begin{minipage}{0.24\textwidth}
    \includegraphics[width=\textwidth]{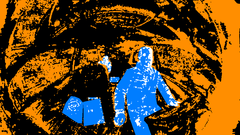}
    \end{minipage}
    \begin{minipage}{0.24\textwidth}
    \includegraphics[width=\textwidth]{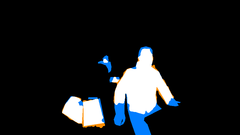}
    \end{minipage}
    \begin{minipage}{0.24\textwidth}
    \includegraphics[width=\textwidth]{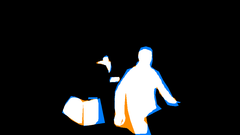}
    \end{minipage}\\[6pt]
    
    \begin{minipage}{0.24\textwidth}
    \includegraphics[width=\textwidth]{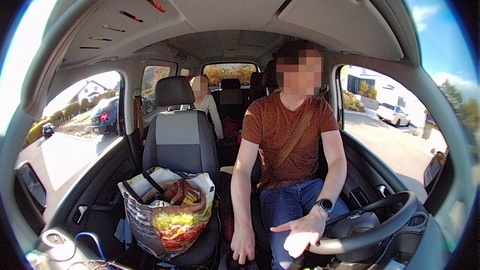}
    \end{minipage}
    \begin{minipage}{0.24\textwidth}
    \includegraphics[width=\textwidth]{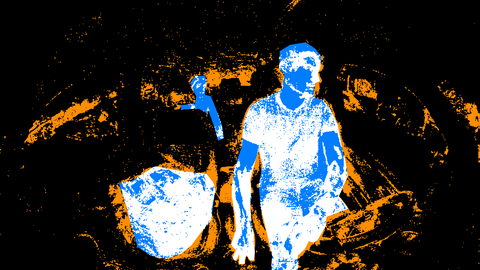}
    \end{minipage}
    \begin{minipage}{0.24\textwidth}
    \includegraphics[width=\textwidth]{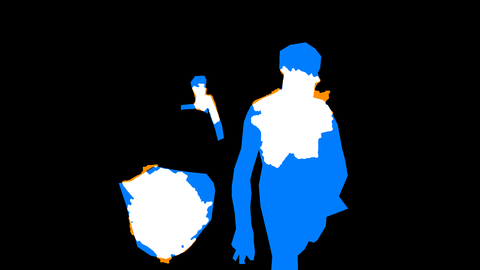}
    \end{minipage}
    \begin{minipage}{0.24\textwidth}
    \includegraphics[width=\textwidth]{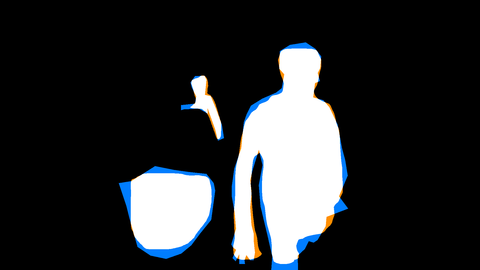}
    \end{minipage}\\[6pt]
    
    \begin{minipage}{0.24\textwidth}
    \includegraphics[width=\textwidth]{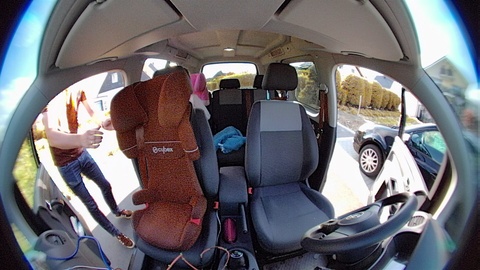}
    \end{minipage}
    \begin{minipage}{0.24\textwidth}
    \includegraphics[width=\textwidth]{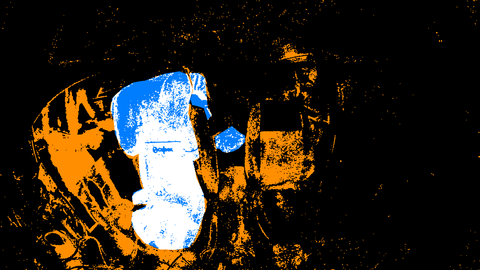}
    \end{minipage}
    \begin{minipage}{0.24\textwidth}
    \includegraphics[width=\textwidth]{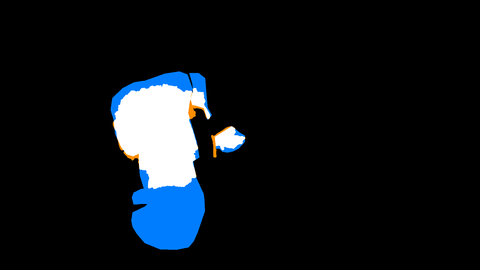}
    \end{minipage}
    \begin{minipage}{0.24\textwidth}
    \includegraphics[width=\textwidth]{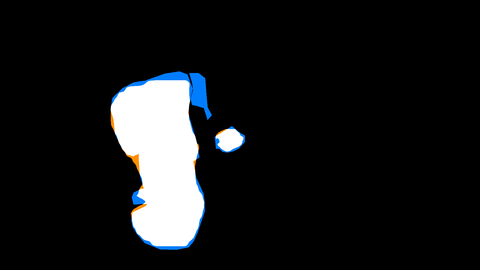}
    \end{minipage}\\[6pt]
    
    \begin{minipage}{0.24\textwidth}
    \includegraphics[width=\textwidth]{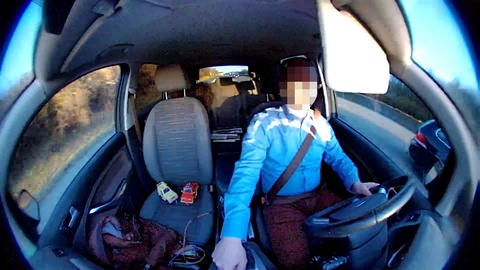}
    \end{minipage}
    \begin{minipage}{0.24\textwidth}
    \includegraphics[width=\textwidth]{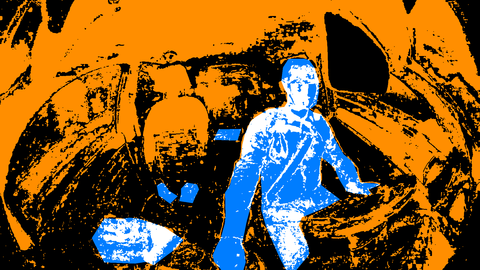}
    \end{minipage}
    \begin{minipage}{0.24\textwidth}
    \includegraphics[width=\textwidth]{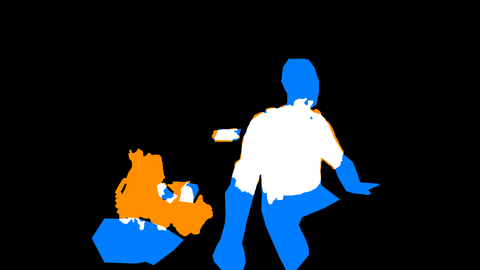}
    \end{minipage}
    \begin{minipage}{0.24\textwidth}
    \includegraphics[width=\textwidth]{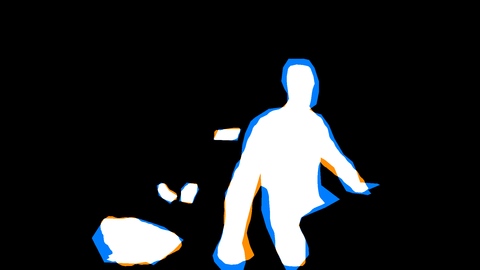}
    \end{minipage}\\[6pt]
    \caption{Comparison of the masks predicted by the investigated methods. 1st column: The RGB real-world images. 2nd column: The masks predicted by the GMM. 3rd column: The masks predicted by the Morphsnakes. 4th column: The masks predicted by the Mask R-CNN.}
    \label{fig:comparison}
\end{figure*}

In summary, the Mask R-CNN clearly outperforms the two classical methods GMM and morphological  snakes. As described in \cref{tab:summary_eval_results} the best model of the Mask R-CNN achieves a similarity score of $75.5\%$ which significantly surpasses the best results of the morphological snakes by $26.9$ pp.\ and of the GMM even by $53.9$ pp. Moreover, we can observe by \cref{fig:comparison} that the Mask R-CNN is the only method which provides a clear segmentation of the foreground instances. By our experiments we also found out that the problems of the classical methods are addressed by the Mask R-CNN. Contrary to the morphological snakes and the GMM, the Mask R-CNN is much more robust against changes in the light conditions, shadows and other exterior factors, like traffic lights.

\begin{table}[!h]
\centering
\resizebox{8.5cm}{!}{
\begin{tabular}{|c|cccccc|}
\hline 
Method & $Pr$ & $Re$ & $Sp$ & $Acc$ & $Sim$ & $F_1$ \\
\hline
Morph snakes & 0.868 & 0.538 & 0.990 & 0.919 & 0.486 & 0.648  \\
\rowcolor{lgray1} 
Mask R-CNN  & 0.919 & 0.797 & 0.991 & 0.969 & {\bf 0.755} & 0.866 \\
GMM &  0.342 & 0.465 & 0.777 & 0.737 & 0.216 & 0.337 \\ 
\hline
\end{tabular}
}
\vspace{1mm}
\caption{Summary of the best evaluation result for each implemented method.}
\label{tab:summary_eval_results}
\end{table}

\section{Conclusion and Outlook}
\label{sec:conclusion}
In this work we have introduced a benchmark for the task of foreground-background segmentation in interior sensing. We compared the segmentation performance of different variants of classical methods, i.e., Gaussian Mixture Models and Morphological Snakes, with the segmentation performance of a recent deep learning model, the Mask R-CNN. Similarly to other real-world computer vision applications, we observe that the Mask R-CNN is much more capable of handling the rather large variety in the recorded scenes. Static and moving objects / persons inside the car as well as static and moving backgrounds outside the car as well as varying illumination and shadows contribute to a complex scenery that classical methods cannot handle anymore. We also found out that the hunger for data of the Mask R-CNN to be reduced to some extent by state of the art data augmentation techniques. However, the only way to sate this hunger seems to be the recording and labeling of new data.

Interesting directions for the future are comparisons with other deep learning models, e.g.\ for semantic segmentation \cite{deeplab}. Besides that, recording additional data that covers difficult cases such as occupied child seats in the back of the car seems of importance. Also deep learning models that consider multiple frames as well as hybrid models that use the output of, e.g., the GMM as an input could be of interest. The latter technique could also be used to reduce the model complexity, making deep neural networks more suitable for embedded systems. Yet, the Mask R-CNN inference requires 0.34 seconds on a Titan XP GPU with 12 GB memory while the GMM model only requires 0.015 seconds per inference on an Intel Xeon E-2186M CPU with 2.90 GHz, a hardware component with way less compute resources.
Furthermore, in the long run, the data collection and labeling process could be supported by 
methods that make proposals towards labeling those scenes that leverage the model's performance the most. This can be approached e.g.\ via active learning for image segmentation \cite{colling2020metabox,AL_SemSeg_ReBased_Bosch}.

\appendices 
\crefalias{section}{appendix}
\section{Excursion: Color Spaces}
\label{ssec:color}
Each pixel is modeled by a color space \cite{dataMining}. We investigate the influence of the choice of the color space for the performance of the implemented methods.  To this end, the considered color spaces RGB, HSV and Lab are briefly introduced below. For the  mathematics behind the conversion between those colors spaces, we refer to \cite{colFundamental} and \cite{colCore}.

\paragraph[RGB]{\bf RGB}
The RGB color space \cite{colFundamental, dataMining, colorSpacesSimulations} can be described as an additive color system since the colors arise by a linear combination of the three primary colors Red, Green and Blue (RGB).
Mathematically, the RGB color space can be comprehended as a cube which is located in a three dimensional Cartesian coordinate system. Thereby, each axis is represented by one of the color channels R, G or B. Thus, RGB colors are given by a three dimensional vector $(r,g,b)$ where $r$, $g$ and $b$ describe the intensities of the corresponding color channels red, green and blue, respectively. The values of the color intensities lie in the interval $[0, M]$. The special case $r=g=b$ represents the colors white $(M,M,M)$, black $(0,0,0)$ and all gray-scale values inbetween.

\begin{figure}[!h]
    \centering
    \includegraphics[trim=0mm 3mm 0mm 3mm, clip, scale=0.6]{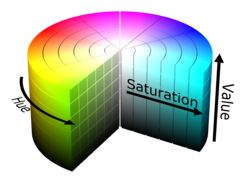}
    \caption{The HSV color space. Source: \cite{hsvGithub}.}
    \label{fig:hsv}
\end{figure}

\paragraph[HSV]{\bf HSV}
The problem about the RGB color space is that the colors are not created according to the color perception of a human. Intuitively, a human creates a color by selecting a color of a certain spectrum and then develop a desired saturation and brightness level. The HSV color space \cite{colFundamental, dataMining, colorSpacesSimulations} is based on this idea. Thus, the goal of the HSV color space is to adapt the definition of colors to the color perception of humans. Hence, the HSV color space can be described as a perceptive color model.
The colors in a HSV space derive by the determination of a Hue, Saturation and 
Value. Mathematically, the HSV color space can be described by a cylinder. The hue is represented by a pure color. All possible pure colors are organized on the borders of the circular base area of the cylinder as described by \cref{fig:hsv}. Hence, the Hue is defined by an angle degree on the base area of the cylinder.
The saturation describes how vibrant a color is and is determined by defining a radius inside the cylinder.
Finally, the color channel V describes the brightness of a color and is represented by a height inside the cylinder. Usually, the value for this channel lies in the range of $[0, M]$ with $0 < M \leq 1$.

\paragraph[Lab]{\bf CIEL$^\mathbf{\ast}$a$^\mathbf{\ast}$b$^\mathbf{\ast}$ (Lab)}
The CIEL$^{\ast}$a$^{\ast}$b$^{\ast}$, also denoted by Lab, was designed by the Commission  Internationale d' Èclairage (CIE) in 1976.
Lab \cite{colCore, lab} is a uniform color space which should correlate with the color perception of humans analogous to the HSV color space.
By the color channel L$^{\ast}$ the luminosity of a color is described. The channels 
a$^{\ast}$ and b$^{\ast}$ represent the color pairs green-red and blue-yellow, respectively. By the value of those color pairs the saturation $ C_{ab}^\ast$ and the hue $h^\ast_{ab}$ of a color are defined as 
\begin{align}
    C_{ab}^\ast &= \sqrt{a^{\ast^2} +b^{\ast^2}} \quad \text{and}\\
    h^\ast_{ab} &= \arctan \left( \frac{b^{\ast}}{a^{\ast} }\right) \, ,
\end{align}
respectively. In this work, especially the color channels V and L are of interest since it is possible to control the brightness of an image by both of these color channels independently of the hue and the saturation. Due to this reason, image enhancement methods regarding the contrast are generally applied on those two color channels. Common contrast enhancement methods, which are also investigated, are given by the Histogram Equalization (HE) \cite{he} and the Contrast Limited Adaptive Histogram Equalization (CLAHE) \cite{clahe}.

\section{Detailed statistics for the ISSO dataset}
\label{sec:appendix_aptiv}
While the statistics over the images of the ISSO test set are given in \cref{tab:test_obj}, the statistics of the ISSO training set are described in \cref{tab:train_object,tab:train,tab:train_person}. Herein, the statistics over the first 500 images of the training set are described by column 
``Original''/``Orig.''.  In column ``Additional''/``Add.'' the statistics over the 600 images by which the training set was extended are captured. The summary of all quantities is given by the column ``Total''. In \cref{tab:train_object} it is also described by the row ``Original'' which objects the first 500 images of the training set contain. In row ``Additional'' the objects are recorded which have been added by the extension of the training set.
\newpage
\begin{table}[!h]
\centering
\begin{tabular}{|l|c|}
\hline
\rowcolor{bblue} Main category 	& Number of instances\\
\hline
					Laptop 			& 4\\
					Jacket 			& 10\\
					Toys			&  6\\
					Trolley			& 1 \\
					Glasses			& 1  \\
					Bag(pack)		&  5  \\
					Stack of books  & 1  \\
					Helmet				&  1 \\
					CD cover 				&   1\\
					Umbrella				& 2  \\
					Smartphone 				&   3\\
					Tissues 				&   1\\
					Wallet 				&  2 \\
					Stack of clothes 				&  1 \\
					Shoe 				&   1\\
					Beverage crate				&  1 \\
					\hline
\end{tabular}
\vspace{1mm}
\caption{Number of instances per main category of the class ``Object'' in the test set.}
\label{tab:test_obj}
\end{table}
\begin{table}[!h]
\centering
\begin{tabular}{|l|l|c|c|c|}
\hline
\rowcolor{bblue}                    & 		& \multicolumn{3}{|c|}{Number of instances}\\ \cline{3-5}
\rowcolor{bblue}	 \multirow{-2}{*}{Images} & \multirow{-2}{*}{Main group} & Orig. & Add. & Total\\
\hline
								& Laptop 							& 3	& 0	& 3	\\
								& Bottle							& 4	& 5	& 9	\\
								& Smartphone	            		& 5 & 0	& 5	\\
								& Blanket	   						& 2 & 0 & 2 \\
								& Jacket    						& 9	& 4	& 13\\
								& Wallet 							& 1	& 0	& 1	\\
								& PC-keyboard 						& 1 & 0	& 1	\\ 
								& Sheet of paper 					& 3	& 2	& 5	\\
								& Bag(pack)							& 10& 10& 20	\\
								& Baker's bag						& 1 & 1	& 2	\\
								& Socks								& 1 & 0	& 1	\\
\multirow{-12}{*}{Original}		& Cardboard box						& 5 & 8	& 13	\\
\hline
								& Suitcase							& 0 & 1	& 1	\\
								& Toys								& 0 & 3	& 3	\\
								& Face mask							& 0 & 1	& 1	\\
								& Hat								& 0 & 2	& 2	\\	
								& Basket							& 0 & 1	& 1	\\
								& Food								& 0	& 2	& 2	\\
\multirow{-7}{*}{Additional}	& Cable drum						& 0	& 1	& 1	\\
\hline
Total							&									&45	&41 & 86	\\
\hline								
\end{tabular}
\vspace{1mm}
\caption{Number of instances per main category of the object class in the training set.}
\label{tab:train_object}
\end{table}
\begin{table}[!h]
\centering
\begin{tabular}{|l|c|c|c|}
\hline
\rowcolor{bblue} 		& \multicolumn{3}{|c|}{Number of instances}\\ \cline{2-4}
\rowcolor{bblue}	 \multirow{-2}{*}{Class}			& Original & Additional & Total\\
\hline
					Person 		& 21& 8& 29\\
					Object 		& 45& 41 &86\\
					Child seat	&  4& 12& 16\\
					\hline
\end{tabular}
\vspace{1mm}
\caption{Number of instances per class in the training set.}
\label{tab:train}
\end{table}
\newpage
\begin{table}[!h]
\centering
\begin{tabular}{|p{1cm}|p{1cm}|c|c|c|}
\hline
\rowcolor{bblue}                    & 		& \multicolumn{3}{|c|}{Number of instances}\\ \cline{3-5}
\rowcolor{bblue}	 \multirow{-2}{*}{} & \multirow{-2}{*}{} & Original & Additional & Total\\
\hline
					& Female 		& 8& 2 &10\\
	\multirow{-2}{*}{Gender}				& Male 		& 13& 6& 19\\
	\hline
					&Baby	    &  1& 1& 2\\
					&Child	    &  0& 1& 1\\
					\multirow{-3}{*}{Age}&Adult	    & 20 & 6& 26\\
					\hline
\end{tabular}
\vspace{1mm}
\caption{Detailed description for the class ``Person'' of the training set regarding the characteristics gender and age.}
\label{tab:train_person}
\end{table}

\section{List of object classes of the COCO dataset}
\label{sec:a1}
In \cref{tab:appendix_coco} the main categories (exclusive the category ``background'') and the corresponding object classes of the COCO training set, year 2017, are described.

\begin{table}[!h]
\centering
\begin{tabular}{|p{1.8cm}|p{5.8cm}|}
\hline
\rowcolor{bblue} Main category & Object classes\\
\hline
person & person\\
\hline
vehicle & bicycle, car, motorcycle, airplane, bus, train, truck, boat\\
\hline
outdoor & traffic light, fire hydrant, stop sign, parking meter, bench\\\hline
animal & bird, cat, dog, horse, sheep, cow, elephant, bear, zebra, giraffe\\\hline
accessory & backpack, umbrella, handbag, tie, suitcase\\\hline
sports & frisbee, skis, snowboard, sports ball, kite, baseball bat,\\ 
		& baseball glove, skateboard, surfboard, tennis racket\\\hline
        kitchen & bottle, wine glass, cup, fork, knife, spoon, bowl\\\hline
        food & banana, apple, sandwich, orange, broccoli, carrot, hot dog,\\
		& pizza, donut, cake\\\hline
furniture & chair, couch, potted plant, bed\\
\hline
\end{tabular}
\vspace{1mm}
\caption{A list of the main categories and their corresponding object categories of the COCO training dataset from 2017.}
\label{tab:appendix_coco}
\end{table}

\section*{Acknowledgment}
%
The authors of this work would like to thank Fabian Kunst from the University of Wuppertal for providing the extended implementation of ``Labelme''. 
H.G. and M.R. acknowledge financial support through the research consortium bergisch.smart.mobility funded by the ministry for economy, innovation, digitalization and energy (MWIDE) of the state North Rhine Westphalia under the grant-no. DMR-1-2.
\includegraphics[scale=0.29]{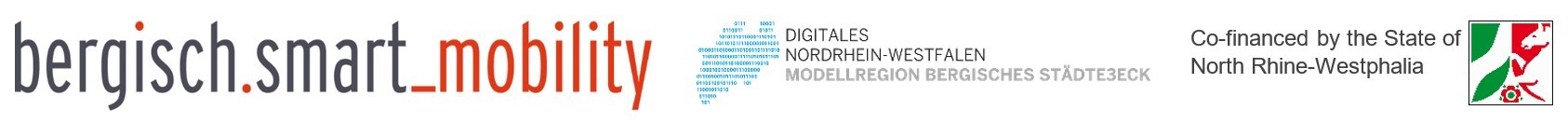}

\ifCLASSOPTIONcaptionsoff
  \newpage
\fi

\printbibliography

\end{document}